\documentclass{article}

\usepackage{arxiv}
\usepackage[utf8]{inputenc} 
\usepackage[T1]{fontenc}    
\usepackage{hyperref}       
\usepackage{url}            
\usepackage{booktabs}       
\usepackage{amsfonts}       
\usepackage{nicefrac}       
\usepackage{microtype}      
\usepackage{lipsum}
\usepackage{fancyhdr}       
\usepackage{graphicx}       
\graphicspath{{media/}}     

\usepackage{graphicx}
\usepackage{amsmath}
\usepackage{amssymb}
\usepackage{booktabs}
\usepackage{comment}
\usepackage{subcaption}
\usepackage{multirow}
\usepackage{arydshln}
\usepackage[capitalize]{cleveref}
\crefname{section}{Sec.}{Secs.}
\Crefname{section}{Section}{Sections}
\Crefname{table}{Table}{Tables}
\crefname{table}{Tab.}{Tabs.}

\newif\ifmodify
 \modifytrue 

\ifmodify
\newcommand{\cross}[1]{\textcolor{red}{\sout{#1}}}

\else
\newcommand{\cross}[1]{}

\fi

\begin{document}

\title{\textbf{DISCO: Distributed Inference with Sparse Communications}}

\author{Minghai Qin, Chao Sun, Jaco Hofmann, Dejan Vucinic\\
Western Digital, Milpitas, California\\
{\tt\small \{minghai.qin, chao.sun, jaco.hofmann, dejan.vucinic\}@wdc.com}
}
\maketitle

\begin{abstract}
Deep neural networks (DNNs) have great potential to solve many real-world problems, but they usually require an extensive amount of computation and memory. 
It is of great difficulty to deploy a large DNN model to a single resource-limited device with small memory capacity.  
Distributed computing is a common approach to reduce single-node memory consumption and to accelerate the inference of DNN models. 
%
%
%
%
%
In this paper, we explore the ``within-layer model parallelism", which distributes the inference of {\em each layer} into multiple nodes. 
In this way, the memory requirement can be distributed to many nodes, making it possible to use several edge devices to infer a large DNN model. 
Due to the dependency within each layer, data communications between nodes during this parallel inference can be a bottleneck when the communication bandwidth is limited.
We propose a framework to train DNN models for Distributed Inference with Sparse Communications (DISCO). We convert the problem of selecting which subset of data to transmit between nodes into a model optimization problem, and derive models with both computation and communication reduction when each layer is inferred on multiple nodes.
We show the benefit of the DISCO framework on a variety of CV tasks such as image classification, object detection, semantic segmentation, and image super resolution. The corresponding models include important DNN building blocks such as convolutions and transformers. 
For example, each layer of a ResNet-50 model can be distributively inferred across two nodes with five times less data communications, almost half overall computations and half memory requirement for a single node, and achieve comparable accuracy to the original ResNet-50 model. This also results in 4.7 times overall inference speedup.

\end{abstract}

\section{Introduction}

Deep neural networks (DNNs) have made great progress in solving real-world problems~\cite{SurveyDNN2021}. A majority of studies assume that DNN models are inferred on a single device, such as a GPU.
There are a few reasons to study the distributed inference of DNN models on multiple devices.
Firstly, good DNN models usually require large memory for inference. This issue becomes more critical as the size of modern DNN models becomes larger and the popularity of resource-limited devices for them to be deployed grows rapidly.
For example, ESP32~\cite{wikiesp32} series of SoCs have only hundreds of kilobytes of RAM/ROM, while a ResNet50 model requires hundreds of megabytes of RAM to store its weights and features, making it extremely challenging to infer a large DNN model on the edge device. By distributing the memory consumption of DNN models to multiple nodes, it is then possible to infer large models by small devices. Second, the inference latency can also be reduced by distributing the computation of inference to multiple nodes as well. Thirdly, some applications can benefit from cooperative inference of several branches DNN models. For example, in the scenario of multi-camera surveillance, if the DNN model in each camera can  not only infer features from its own view, but collaboratively receive/send data from/to other cameras, the detection and classification of an object or an action can be more accurate. The overall system can be viewed as a distributed inference of a larger DNN model with communicated features between nodes.

When the memory capacity is satisfied by distributed hardware, two of the most popular metrics to measure the performance of a DNN-based system are the latency and the throughput~\cite{Hanhirova2018}. The latency is the time difference between a data sample is fed into the DNN and the output is obtained. The throughput is the number of data samples the system can process per time unit.
While throughput plays an important role in training, the latency in inference is a critical criterion of the system for time-sensitive applications, such as autonomous driving~\cite{grigorescu2020survey}, tele-surgery~\cite{fekri2018towards},  security cameras~\cite{pang2021deep}, and so on.
Therefore, the inference latency is our primary interest in this paper.

Parallelism by distributing the computation across many nodes is one of the most popular methods to accelerate the DNN and it can also reduce the memory requirement for each node~\cite{ParallelismNIPS2012,LI201695}.
In the {\em data parallelism}, each node is responsible for processing a subset of input samples in one mini-batch.
In the {\em pipeline parallelism}~\cite{JointDNN2021, Offloading2018, Neurosurgeon2017, CRIME2021}, the DNN model is partitioned into sequential parts based on the execution order, and each node is responsible for processing one part. The DNN is therefore processed in the pipeline manner.
However, neither data nor pipeline parallelism can reduce the latency during the inference of one input. This is because the input data has to be sequentially processed through all DNN layers, and within each layer the computation is not distributed.

\begin{figure*}[t]
    \centering
    \includegraphics[width=1\linewidth]{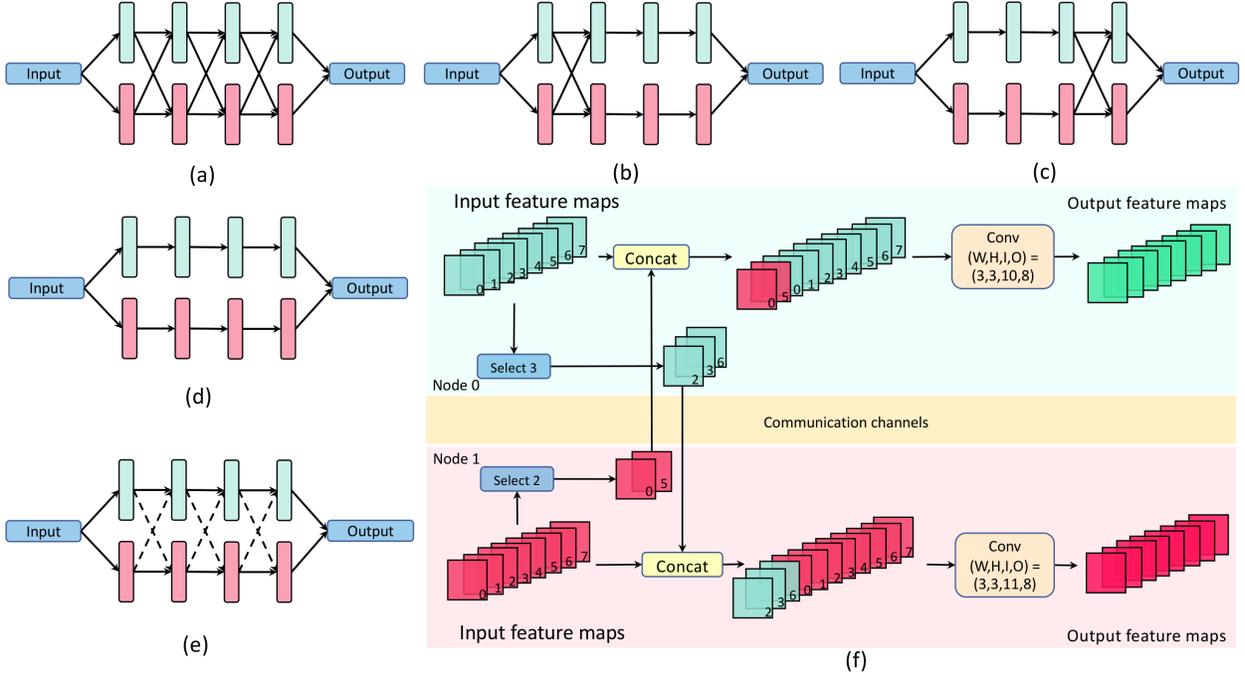}
    \small
    \caption{Comparison of different distributed DNN inference methods. (a) Conventional dense communications. (b) Dense-then-split~(\cite{kim2017icmlsplitnet}). (c) Split-then-aggregate (\cite{dong2022cvprsplitnets}). (d) Independent branches~(\cite{Hadidi2020LCP}). (e) Distributed inference with sparse communications (DISCO, ours). (f) Details of one layer in an example 2-node DISCO. Each node has 8 features from previous layers. Node-0 and Node-1 select 3 and 2 data features (out of 8) to transfer to the other node, respectively. The transferred features are combined with its own features to form the input to the convolutional kernels. The output of the convolution (after point-wise non-linear activation functions) will again be served as the input to the next layer with sparse communications. }
    \label{fig:disco}

\end{figure*}

In order to distribute the computation within a layer, the {\em model parallelism} is proposed where each node is responsible for computing a subset of the output of a layer~\cite{LI201695}. However, each feature in the output of a layer is usually dependent on \textbf{all} input data of the layer, such that all input data needs to be distributed to all nodes (Figure~\ref{fig:disco}(a)).
The data communication latency between nodes will be the bottleneck of the distributed system with low communication bandwidth.
For example, based on the configuration of a head-mounted system in~\cite{dong2022cvprsplitnets}, the communication latency can be one or two orders of magnitudes higher than the computation latency (See Figure~\ref{fig:resnet50_layer_profile} in Section~\ref{sec:Num_input_features_to_comm}) for layers of a ResNet50 model.
This phenomenon also happens for high-end (e.g., A100 GPUs and NV-Links) and low-end (e.g., ARM cores and wireless links) platforms.
Many prior works~\cite{HuDiyi2020, Jung-Lee2021,DEFER2022,Deeperthings2021,DeepThings2018} tried to use better scheduling methods to improve the overall system performance.
However, the potential of jointly designing the DNN architectures based on the system configurations is not explored.
Another trend to distribute the computation is by splitting some stages of the DNN into independent branches~\cite{kim2017icmlsplitnet, dong2022cvprsplitnets,Hadidi2020LCP} (Figure~\ref{fig:disco}(b)(c)(d)). In these works, the separate branches can be processed independently on different nodes and there is no data communication between them. This eliminates the data communication latency for the corresponding stages. However, completely separated branches hurt the accuracy.

In this paper, we jointly design the DNN architecture and system configurations for distributed inference.
In our proposal, only a subset of input features are transmitted between nodes (Figure~\ref{fig:disco}(e)). The problem of selecting an optimal subset of them to communicate between nodes is NP-hard.
We first demonstrate that this problem is equivalent to the problem of DNN weight pruning with certain patterns.
To be more specific, weights in a convolutional or fully-connected layer can be represented as a 2-dimensional matrix $M$. And if Node-$i$ does not need a specific input feature in Node-$j$ to compute the output features in Node-$i$, it is equivalent to setting a sub-row of appropriate size and location in $M$ to be all-zero. Details are presented in Section~\ref{sec:selection}.
In our framework, we first train a complete model with dense communication. Then we identify the non-zeros weights corresponding to the sparse features to communicate. We gradually sparsify the communication by fine-tuning the remaining weights.
This framework, called {\em distributed inference with sparse communications (DISCO)}, can search for models with a better trade-off between data communication latency, computation latency, and accuracy.
In addition, observing that the data communications and computation can be pipelined to hide their latency, the proposed method can further reduce the end-to-end system latency at almost no cost to accuracy.
By experimenting on multiple machine learning tasks, including image recognition, object detection, semantic segmentation, and super-resolution,
we conclude that 1) Compared to DNNs with completely independent branches, the accuracy can be significantly improved by a very tiny proportion of the data communications; 2) Compared to DNNs with completely dependent input-output layers (dense data communications), our proposed method can significantly reduce data communications with negligible accuracy loss.




The contribution of this paper is summarized as follows.

\textbf{1)}. We propose a framework, called {\em DISCO}, to design distributed DNN model architectures with sparse communications based on system configurations.

\textbf{2)}. We proposed to convert the problem of selecting the subset of data to communicate (which is NP hard) to a DNN weight pruning problem with certain structured patterns.


\textbf{3)}. The proposed DISCO can reduce the average inference latency by 4.4x based on system configurations in~\cite{dong2022cvprsplitnets}.
We also calculate the latency improvement based on four publicly available system specifications and observed 3.5x average speedup without accuracy degradation. The results are based on the geometric mean of five types of DNN architectures and four system configurations (See Table~\ref{tab:summary_pipeline}).
Compared with prior arts, the DISCO can increase accuracy by 1.6\% on image classification, 2.8 mAP on object detection, 6.6 mIoU on semantic segmentation, and 0.6 dB PSNR on super-resolution.
As far as we know, this covers the widest range of applications compared to similar works.

\section{Related Works}

\subsection{Distributed inference}
\subsubsection{Network splitting}
Some prior works assume a device-cloud scenario where the layers of neural network are distributed among them~\cite{DDDC2017,dong2022cvprsplitnets,MCDNN2016,DistilledSplit2019,Mohammed2020}. Many works focus on selecting the splitting point for separate device and cloud execution~\cite{JointDNN2021, Offloading2018, Neurosurgeon2017, CRIME2021}. In these methods, the communication happens at the splitting point and no other communication latency is considered within each side. Our work, on the contrary, focuses on the fundamentals of distributed inference where no difference in the computational capabilities of processors is assumed and we consider the communication latency of \textbf{all} layers. Therefore, our work can also be used to improve their latency on either the device or the cloud side.

~\cite{DISSEC2022,kim2017icmlsplitnet,huchenghao2022,Hadidi2020LCP} partition the neural network into several branches and each branch is distributed to a node for independent execution. Our work is different in that our network design is not completely separate such that there is communication between nodes. We can show that a tiny fraction of communication (e.g., 1\%) can improve accuracy significantly with our framework.

\subsubsection{Network scheduling}
Unlike works with changes to DNNs, there are some other works focusing on the system scheduling of inference without changing the models.~\cite{HuDiyi2020} uses compression and dynamic scheduling to improve the throughput.~\cite{Jung-Lee2021} forms the splitting problem into a graph routing problem.~\cite{DEFER2022} presents a framework to dispatch the inference data to compute nodes. ~\cite{DeepThings2018,Deeperthings2021} minimizes the footprint in parallelism and enables dynamic workload distribution.~\cite{DistrEdge2022} considers heterogeneous and non-linear device characteristics and proposes splitting and distributing methods. Compared to system design methods where the focus is to find a good schedule for unchanged models, our work is different, and can be superimposed on theirs as well, in that we modify the network architecture to directly reduce data communications to achieve a better accuracy-latency trade-off. 


\subsection{Distributed training}
Model and pipeline parallelism can accelerate the distributed training. Graph partitioning~\cite{AMPNet2017,Giacomoni2008,Gordon2006ExploitingCT,Guan2019XPipeEP,PipeDream2018,ChenYangCheng2018,Tanaka2021AutomaticGP} splits the training data into micro-batches and they are pipelined into the training devices. ~\cite{Megatron2019,JiaLin2018,WangHuangLi2019} split the tensors and place them onto different devices to distribute the computing. 
Data parallelism encounters the problem of high communication bandwidth for transmitting gradients.~\cite{AbrahamyanChenBekoulis2021,Aji2017SparseCF,Strom2015,QSGD2017,NestedDQ2019} propose methods to sparsify and quantize the gradients to reduce communication, respectively.

While the distributed training research focuses on improving the training throughput, our work, on the other hand, focuses on improving the \textbf{distributed inference latency} of each DNN layer by a slight architectural change.
\section{Methodology}


\begin{figure*}[t]
    \centering
    \includegraphics[width=1\textwidth]{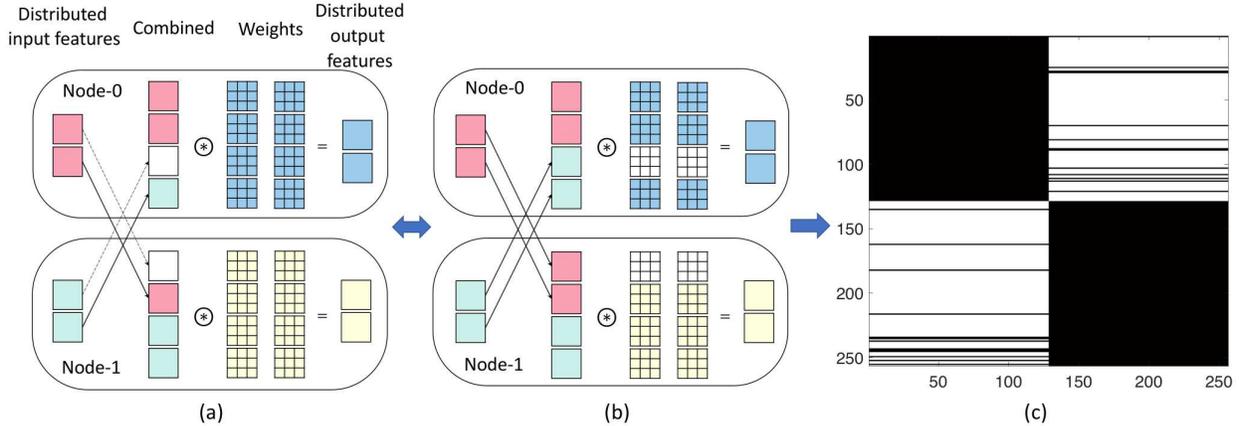}
    \small
    \caption{Equivalence of feature selection and weight selection. (a) One of two features is transmitted from Node-0 to Node-1 and from Node-1 to Node-0, respectively. The non-transmitted feature can be viewed as all-zero (a blank feature). (b) The equivalent representation where all features are transmitted but the corresponding weights are all-zero (blank 3-by-3 kernels). (c) A real example of the weight matrix in a ResNet-50 model found by our DISCO. The number of both input and output feature maps are 256. Each black subrow in the anti-diagonal represents one transmitted input feature between the two nodes.}
    \label{fig:equivalance}
\end{figure*}

\subsection{Problem statement} \label{sec:problem_statement}

The basic operation of the DNN model inference is the multiplication of input features and weights. The input features are feature maps for convolutional neural networks (CNNs) and are neuron vectors/matrices for multi-layer perceptions (MLPs) and Transformers.
Assume there are $I$ input features. In order to distribute the computation of each DNN layer into $N$ nodes, they are equally distributed with $\frac{I}{N}$ input features per node.  If an output feature in Node-$i$ is dependent on input features in Node-$j$, the corresponding dependent input features are required to be transmitted from Node-$j$ to Node-$i$, inducing the communications from Node-$j$ to Node-$i$. Figure~\ref{fig:disco}(f) demonstrates an example of distributing a convolutional layer with $3\times 3$ kernels into $N=2$ nodes. The total number of input features is $I=16$ and each node carries $8$ of them. For conventional DNNs with dense communications shown in Figure~\ref{fig:disco}(a), the convolution tensor on each node has the shape $(W,H,I,O)=(3,3,16,8)$, where $W$ and $H$ are the kernel size, $I$ and $O$ are 16 input features and 8 output features, respectively. The total number of output feature maps is then $8+8=16$. Though the computation of the convolution operation is distributed across 2 nodes, the communication latency may be a bottleneck of the system since all 8 input features need to be transmitted across the communication channel in both ways.

In DISCO, in order to reduce the communication latency, on the other hand, for example, Node-0 can select 3 input features, which are the 2nd, 3rd, and 6th feature maps, and sends them to Node-1, resulting in $\frac{5}{8}$ one-way communication reduction. Similarly, in the other direction, the selection of the 0th and 5th feature maps from Node-1 to Node-0 can reduce the data transmission by $\frac{6}{8}$. The problem of designing the DNN model for distributed inference can be formulated as the following few questions:
\begin{enumerate}
\item
If the total number of features to be communicated is known, which subset of input features should be selected to transmit to other nodes?
\item
If the subset of transmitted input features is determined, how should the weights of the DNN be trained?
\item
How many input features should be communicated between nodes?
\end{enumerate}

\subsection{Selection of the subset of input features to communicate}\label{sec:selection}
In this section, we use a convolutional layer to demonstrate the subset selection problem and it can be generalized to MLP and Transformers as well.

Let $I_{\textrm{each}} = \frac{I}{N}$ be the number of local input features on each node, and assume the number of input features communicated from Node-$i$ to Node-$j$ is $I_{\textrm{comm}}$, then there are $\binom{I_{\textrm{each}}}{I_{\textrm{comm}}}$ possible subsets of input features to select. This problem is NP-hard for moderate fraction $\frac{I_{\textrm{comm}}}{I_{\textrm{each}}}$. For example, if $I_{\textrm{each}}=128$ and $I_{\textrm{comm}}=32$ (transmitting a quarter of features), then the number of subset selections is larger than $10^{30}$, making it impossible to enumerate.

Let the subset of input features that are {\em not} transmitted to the other node be denoted by $\mathcal F$, it is mathematically equivalent to transmitting them as all-zero features $\mathcal F_{\textrm{zero}}$ and the corresponding weights $W_{\mathcal F}$ convolving these all-zero features can be arbitrary (Figure~\ref{fig:equivalance}(a)). This is because  zero times any value equals zero. Then, it is mathematically equivalent to transmit all input features including $\mathcal F$ and set $W_{\mathcal F}=\boldsymbol{0}$, as in Figure~\ref{fig:equivalance}(b). Therefore, the problem of selecting a subset of features, which is dynamic and related to the dataset distribution, can be converted into a weight pruning problem, where the pruning pattern is a sequence of convolution kernels (e.g., a sequence of $3 \times 3$ kernels) that corresponds to input features in the other nodes. Therefore, if the 4-D convolution weight tensor is projected to a 2-D mask-matrix by condensing each 2-D convolution kernel into one value representing whether this kernel is all-zero, then the pruned pattern is a half row of the matrix (for 2 nodes), or in general, a $\frac{1}{N}$-th row of the matrix, because each sub-row represents the weights that convolve one input feature in the other nodes.
Note that for Node-$i$, the convolution kernels corresponding to its input features are always kept. Therefore, the diagonal blocks of the mask-matrix are always kept. Figure~\ref{fig:equivalance}(c) shows the 2-node mask-matrix for one of the layers in ResNet-50 that has 256 input and output feature maps, respectively. The communication sparsity is 90\%. Each point in the mask represents a $3\times 3$ convolutional kernel. Two dark blocks on the diagonal represent the dense weights and communications within each node. In the anti-diagonal, white regions mean the corresponding weights are pruned and the dark black sub-rows correspond to those input features that will be communicated between the two nodes.

After converting the input feature selection problem to a weight pruning problem, we can find an approximation of the problem efficiently by estimating the significance of the weights. First, a full DNN model with dense communications is trained.
Second, for each layer, create a mask matrix $M\in \{0,1\}^{I \times O}$ for each convolutional kernel of shape $W\times H$ (e.g., $3\times 3$ or $1\times 1$), where $I$ and $O$ are the number of input and output features, respectively.
Then, for the non-diagonal sub-row of size $1$-by-$\frac{O}{N}$ that corresponds to $\frac{OWH}{N}$ weights in the full model, calculate the $L_1$ norm of them and prune the sub-rows  with least $L_1$ norms to all-zero values.
If we partition mask $M$ into $\frac{I}{N} \times \frac{O}{N}$ sub-matrices, then pruning one non-diagonal sub-row in the $(i,j)$-th sub-matrix corresponds to preventing one input feature from being transmitted from Node-$i$ to Node-$j$.
The total number of pruned sub-rows depends on the desired sparsity in the non-diagonal matrix, which is the communication sparsity of this layer.
The remaining sub-rows are the input features with greater significance and should be communicated between nodes for optimum latency-accuracy trade-offs. We ablate over the random sub-row pruning to show the benefit of DISCO.

\subsection{Training DNNs with sparse communications} \label{sec:iterative_training}
After setting up the equivalence between sparse weights and sparse communications, we use the simple yet effective iterative magnitude prune-and-finetune procedure to obtain trained DNNs with sparse communications. First, we train a full model with dense communications. Then we prune the least significant sub-rows in the weight matrix (see Section~\ref{sec:selection}) by a fraction of $p_1$, and then finetune the remaining weights to recover some accuracy. This completes one iteration. We perform this fraction-$p_i$ prune and finetune process for a few iterations to generate a sequence of models with increasing weight sparsity, which corresponds to models with increasing communication and computation sparsity. The relationships between weight, communication, and computation sparsity will be discussed in Section~\ref{sec:Num_input_features_to_comm}.
If we set the target communication sparsity to be 100\% in DISCO, then the resultant model has the same architecture as a model with all independent branches in Figure~\ref{fig:disco}(d). Nonetheless, the model obtained by  DISCO significantly outperforms directly training models with all independent branches, under the same number of training efforts (e.g., epochs).


\subsection{Number of input features to communicate} \label{sec:Num_input_features_to_comm}
The total latency of the system depends on the latency of computation and communications. Assume a conventional DNN layer  has computation complexity $C_c$ (in the number of floating-point operations, FLOPs) and the total size of features being $F_s$ (in Bytes). The inference of the layer is distributed across $N$ nodes.
If there is no sparsity in communications, each node will send out its own input features of size $\frac{F_s}{N}$ and receive all other features from other nodes of total size $\frac{(N-1)F_s}{N}$. Therefore, the total data communication of each node is $\frac{F_s}{N}+\frac{(N-1)F_s}{N}=F_s.$
Note that if the speed of sending and receiving is different, we can calculate the {\em weighted} sum of $\frac{F_s}{N}$ and $\frac{(N-1)F_s}{N}$. To avoid complication of adding extra system parameters, we assume equal sending/receiving communication bandwidth, denoted by $B$ (in Bytes/second).
Let $S_{\textrm{comm}}$ denote the communication sparsity of each node, then the latency of the communications is
\begin{equation} \label{eq:Lcomm}
    L_{\textrm{comm}} = \frac{F_s(1-S_\textrm{comm})}{B}.
\end{equation}

Let $S_{\textrm{comp}}$ denote the computation sparsity of the {\em entire} layer (which also equals to the weight sparsity $S_{\textrm{wt}}$), and let $C$ denote the computation speed (in operations/second), then the latency of the computation for each node is
\begin{equation} \label{eq:Lcomp}
L_{\textrm{comp}} = \frac{C_c(1-S_{\textrm{comp}})}{NC}.
\end{equation}

\begin{figure}[t]
    \centering
    \includegraphics[width=1\linewidth]{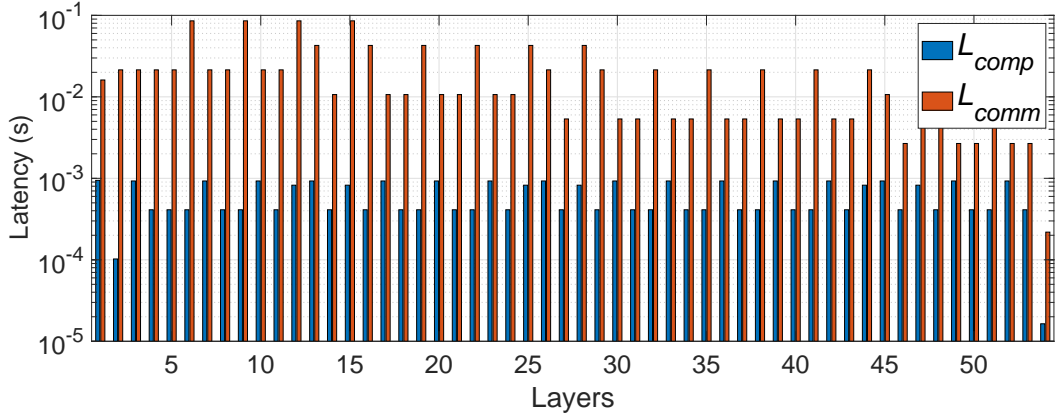}
    \small
    \caption{Computation and communication latency of 54 layers (including 4 skip links) in a ResNet-50 with dense communications between $N=2$ nodes. The system configuration is in~\cite{dong2022cvprsplitnets}, with $B=37.5$ MB/s, $C=125$ GOP/s. }
    \label{fig:resnet50_layer_profile}
\end{figure}

Figure~\ref{fig:resnet50_layer_profile} shows $L_{\textrm{comm}}$ and $L_{\textrm{comp}}$ for each layer of a ResNet-50 model with dense communications ($S_{\textrm{comp}}=S_{\textrm{comm}}=0$) when the inference is distributed to $N=2$ nodes. We assume each value in the feature maps is represented as a 4-byte floating-point number. The system configuration is assumed to be the same in~\cite{dong2022cvprsplitnets} where $C={125}$ GOP/s and $B=37.5$ MB/s. Note that the Y-axis is in the logarithmic scale and the communication latency can be larger than the computation latency by one or two orders of magnitude. This phenomenon can also be observed on other typical computing platforms, from high-end GPUs (NVIDIA A100 with NV-Link) to low-end edge devices (Arm Cortex-M4 with wireless connections). Therefore, reducing communication latency can significantly improve the overall inference performance.

From Figure~\ref{fig:disco}(f), we can observe that sparse communication can also bring computation reduction. The relationship can be described as
\begin{equation} \label{eq:scomm-and-scomp}
S_{\textrm{comm}} = \frac{N}{N-1} S_{\textrm{comp}} =  \frac{N}{N-1} S_{\textrm{wt}}.
\end{equation}
The proof is given in Appendix.

\subsubsection{Communications and computation pipelining}
Prior work, e.g.,~\cite{dong2022cvprsplitnets}, calculates the end-to-end system inference latency by adding the latency of communication and computation.
In this work, on the other hand, we note that computation consists of a small portion of the communicated features and a large portion of local features. Thus, asynchronous computation and communication can be pipelined to ``hide'' the latency of each other.
For example, one node can simultaneously start processing the convolution of features and weights of its own and receiving features from another node.
Therefore, in the ``pipeline'' mode, the latency of inferring Layer-$i$ is the maximum latency of computation or communication, i.e.,
\begin{align} \label{eq:pipeline}
    L(i) = \max \{ L_{\textrm{comm}}(i), L_{\textrm{comp}}(i) \},
\end{align}
where $L_{\textrm{comm}}(i)$ and $L_{\textrm{comp}}(i)$ denote the corresponding latency of Layer-$i$ and they can be computed by Eq.~\ref{eq:Lcomm} and Eq.~\ref{eq:Lcomp}. On the other hand, ``waiting'' mode means that the compute starts after feature communication is completed  ($L(i) = L_{\textrm{comm}}(i)  + L_{\textrm{comp}}(i) $). The overall latency of the DNN is then the summation of latencies of all layers.
We present the results of the ``pipeline'' mode in this main context to demonstrate our performance improvement. The improvements of our method for the ``waiting'' mode are similar.

\subsubsection{Computation and communication equilibriums} \label{sec:equilibrium}
\begin{figure}[t]
    \centering
    \includegraphics[width=1\linewidth]{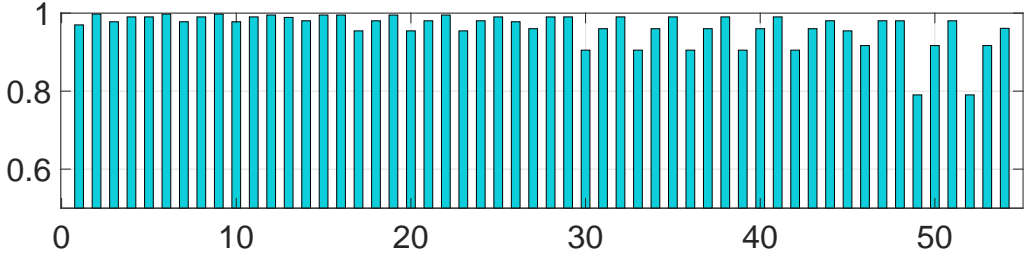}
    \small
    \caption{Non-uniform communication sparsity of each  layer in a ResNet-50 model such that $L_{\textrm{comm}}(i)=L_{\textrm{comp}}(i)$. }
    \label{fig:result_rn50_equilibrium}

\end{figure}

From Eq.~\ref{eq:pipeline}, the latency of a layer is determined by the slower one between $L_{\textrm{comm}}$ and $L_{\textrm{comp}}$. From Eq.~\ref{eq:Lcomm}, Eq.~\ref{eq:Lcomp}, and Eq.~\ref{eq:scomm-and-scomp}, both $L_{\textrm{comm}}$ and $L_{\textrm{comp}}$ will be reduced by increasing $S_\textrm{comm}$. Therefore, there exists an equilibrium point $S_\textrm{comm}^{\textrm{eql}}$ where $L_{\textrm{comm}}=L_{\textrm{comp}}$. If $S_\textrm{comm} < S_\textrm{comm}^{\textrm{eql}}$, the latency reduction of the layer is contributed by reducing $L_{\textrm{comm}}$; On the other hand, if $S_\textrm{comm} > S_\textrm{comm}^{\textrm{eql}}$, further reducing $L_{\textrm{comm}}$ will induce less effective accuracy-latency trade-off than before because  $L_{\textrm{comp}}$ is reduced slower than $L_{\textrm{comm}}$ beyond the equilibrium point and it dominates the latency of the layer.
By setting $L_{\textrm{comm}}=L_{\textrm{comp}}$ in Eq.~\ref{eq:Lcomm},~\ref{eq:Lcomp},~\ref{eq:scomm-and-scomp}, we can obtain the equilibrium sparsity of communication for each layer as
\begin{align}
    S_\textrm{comm}^{\textrm{eql}} = \frac{AN-N}{AN-N+1},
\end{align}
where $A = \frac{N C}{C_c}\cdot \frac{F_s}{B}.$ Note that $S_\textrm{comm}^{\textrm{eql}}\in [0,1]$ if $A\geq 1$, that is, the system bottleneck is the communication latency.
Figure~\ref{fig:result_rn50_equilibrium} shows the $S_\textrm{comm}$ at the equilibrium for the system configuration presented in Figure~\ref{fig:resnet50_layer_profile}.
It can be observed that earlier layers favor higher communication sparsity than later layers.
We also observe that for the ResNet-50 architecture, it is helpful to maintain the accuracy if later layers have smaller sparsity, which coincides with the equilibrium point. Therefore, DISCO at the equilibrium is more likely to achieve a better accuracy-latency trade-off. This is confirmed by the ablation of the communication sparsity in the ResNet-50 experiments.

\section{Experimental Results}
In this section, we demonstrate the benefits of DISCO through several machine learning tasks, including image classification (the ResNet and DeiT models), object detection (the SSD models), semantic segmentation (the DeepLabV3+ models), super resolution (the ESRGAN models), machine translation (the encoder-decoder Transformer models). All models are trained on NVIDIA A100 GPUs by PyTorch implementation.

\begin{figure*}[t]
\centering
\begin{minipage}[c]{0.45\linewidth}
  \hspace{-1em}
  \includegraphics[width=1\linewidth]{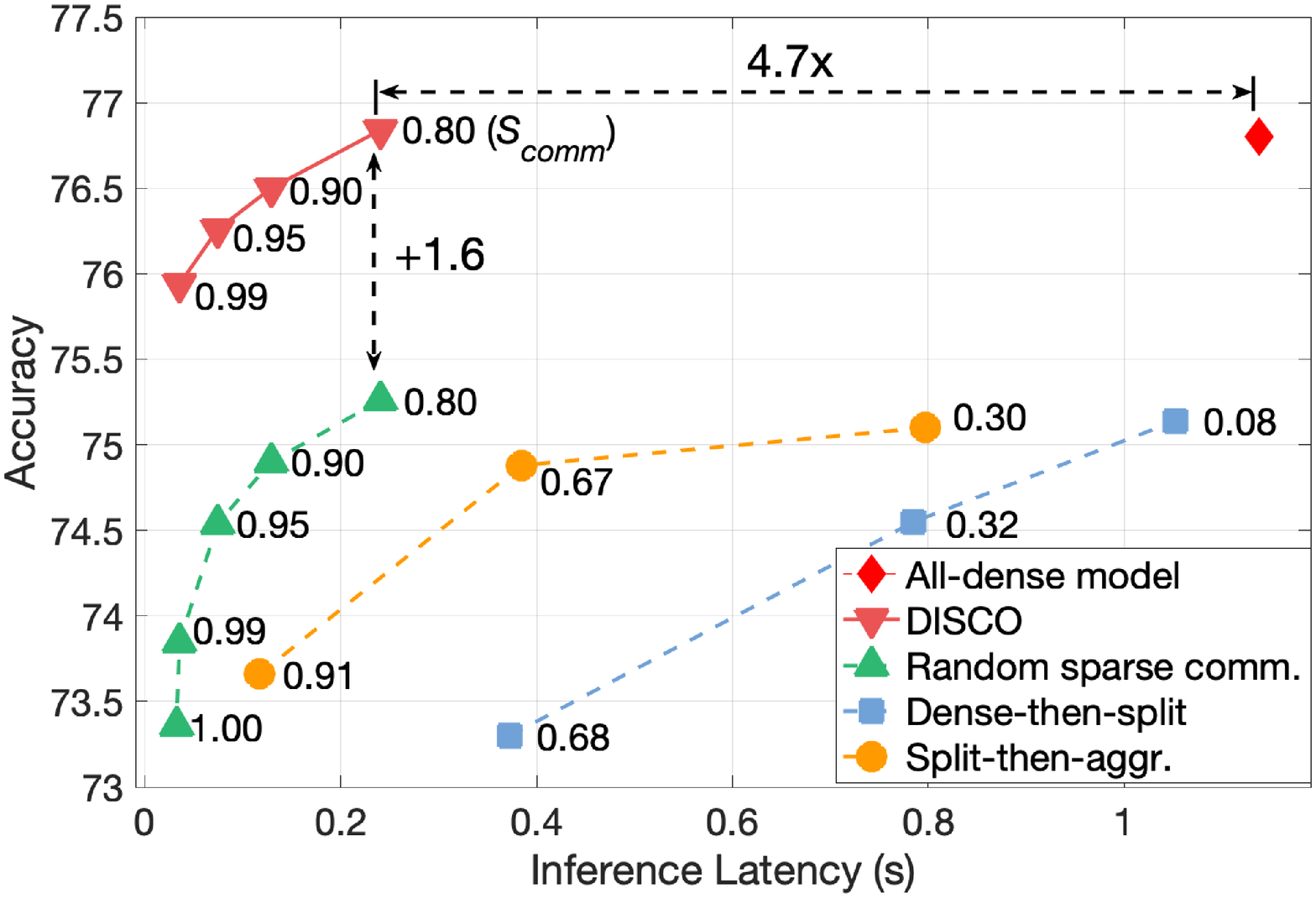}
  \caption{ResNet-50, ImageNet, Top-1}
  \label{fig:result_rn50}
\end{minipage} 
\begin{minipage}[c]{0.45\linewidth}
    \hspace{-1em}
  \includegraphics[width=1\linewidth]{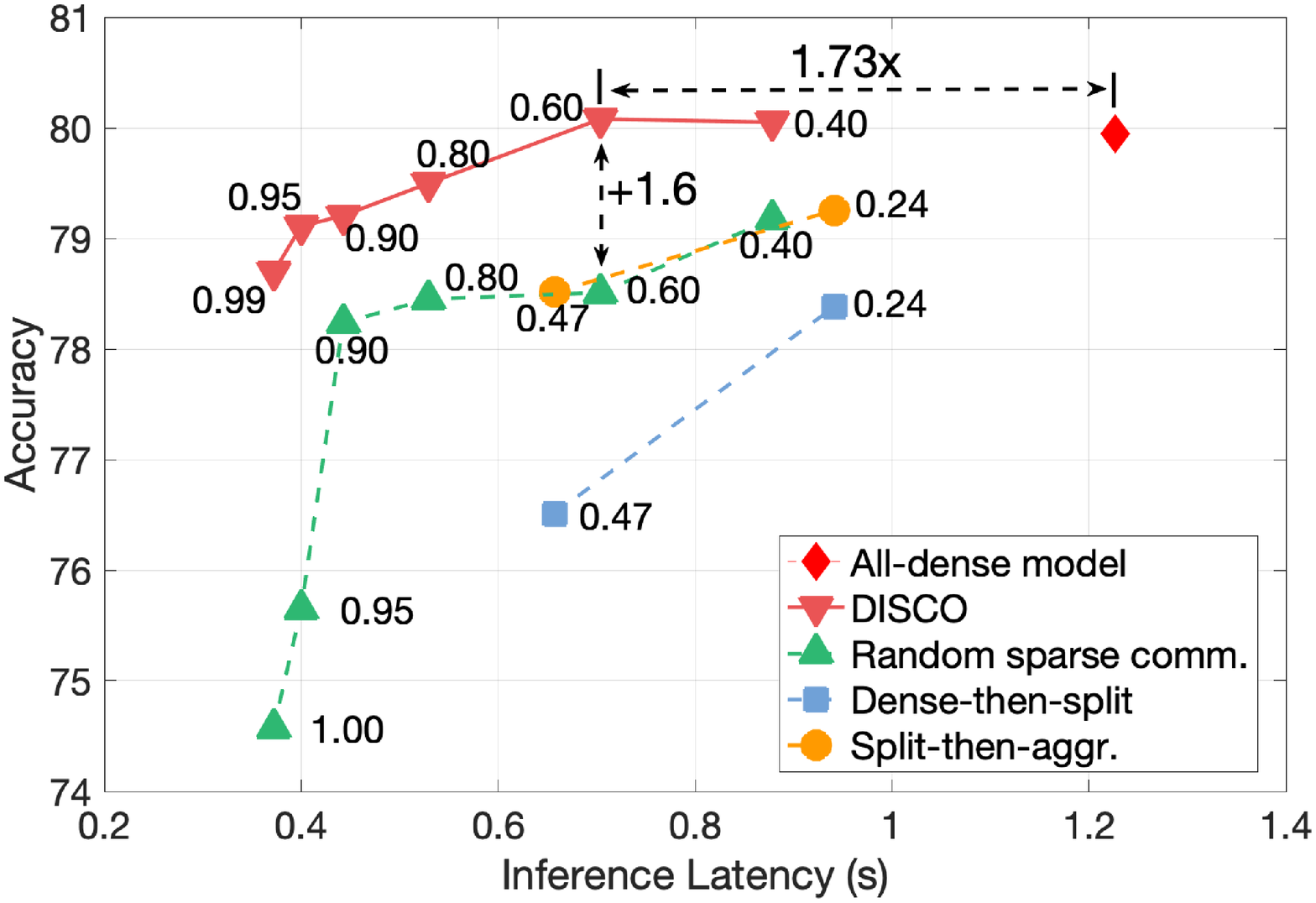}
  \caption{DeiT-S,  ImageNet, Top-1}
  \label{fig:deit}
\end{minipage}

\begin{minipage}[c]{0.45\linewidth}
  \hspace{-1em}
  \includegraphics[width=1\linewidth]{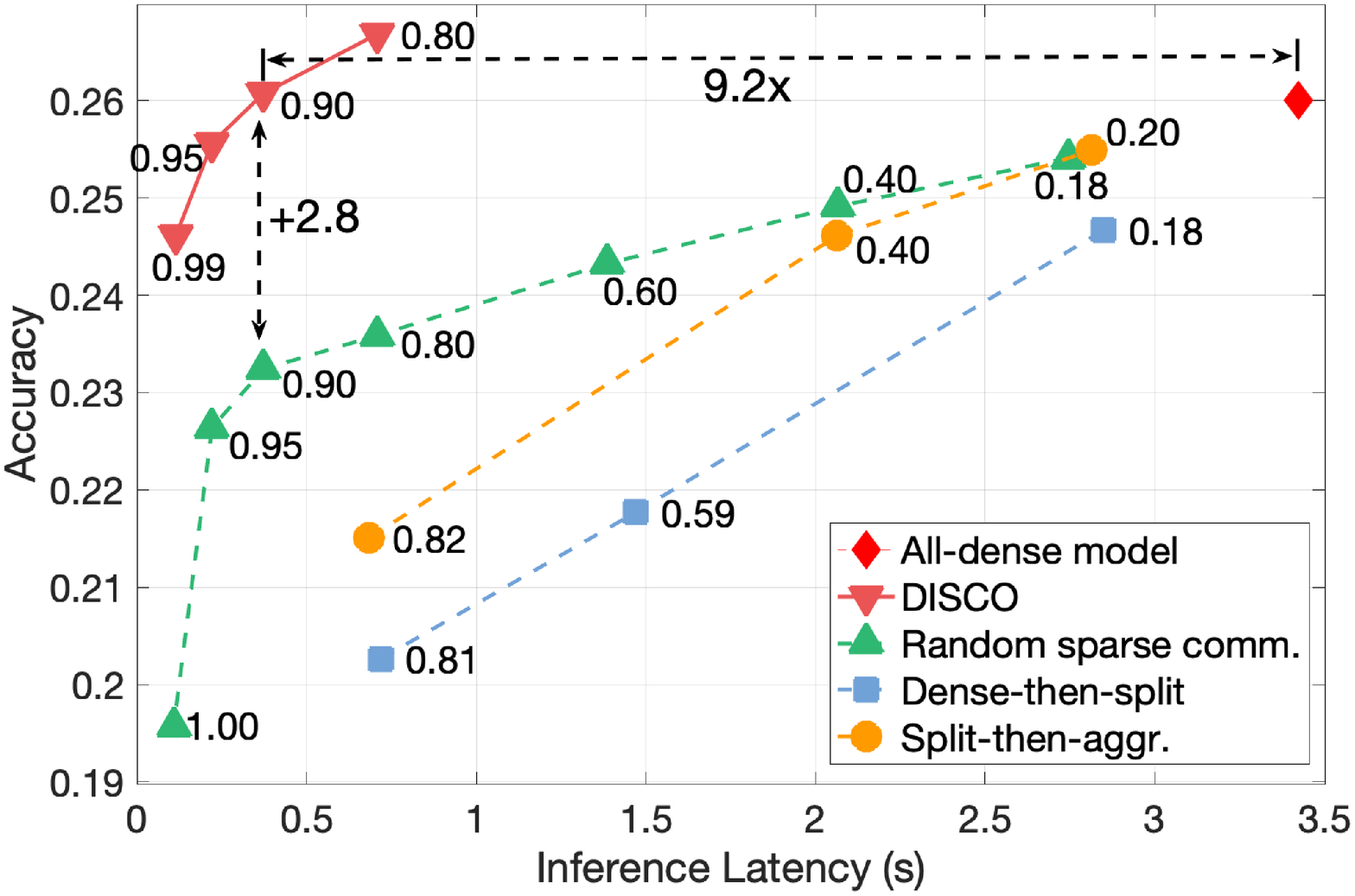}
  \caption{SSD, COCO2017, mAP}
  \label{fig:result_ssd300}
\end{minipage}
\begin{minipage}[c]{0.45\linewidth}
\hspace{-1em}   
  \includegraphics[width=1\linewidth]{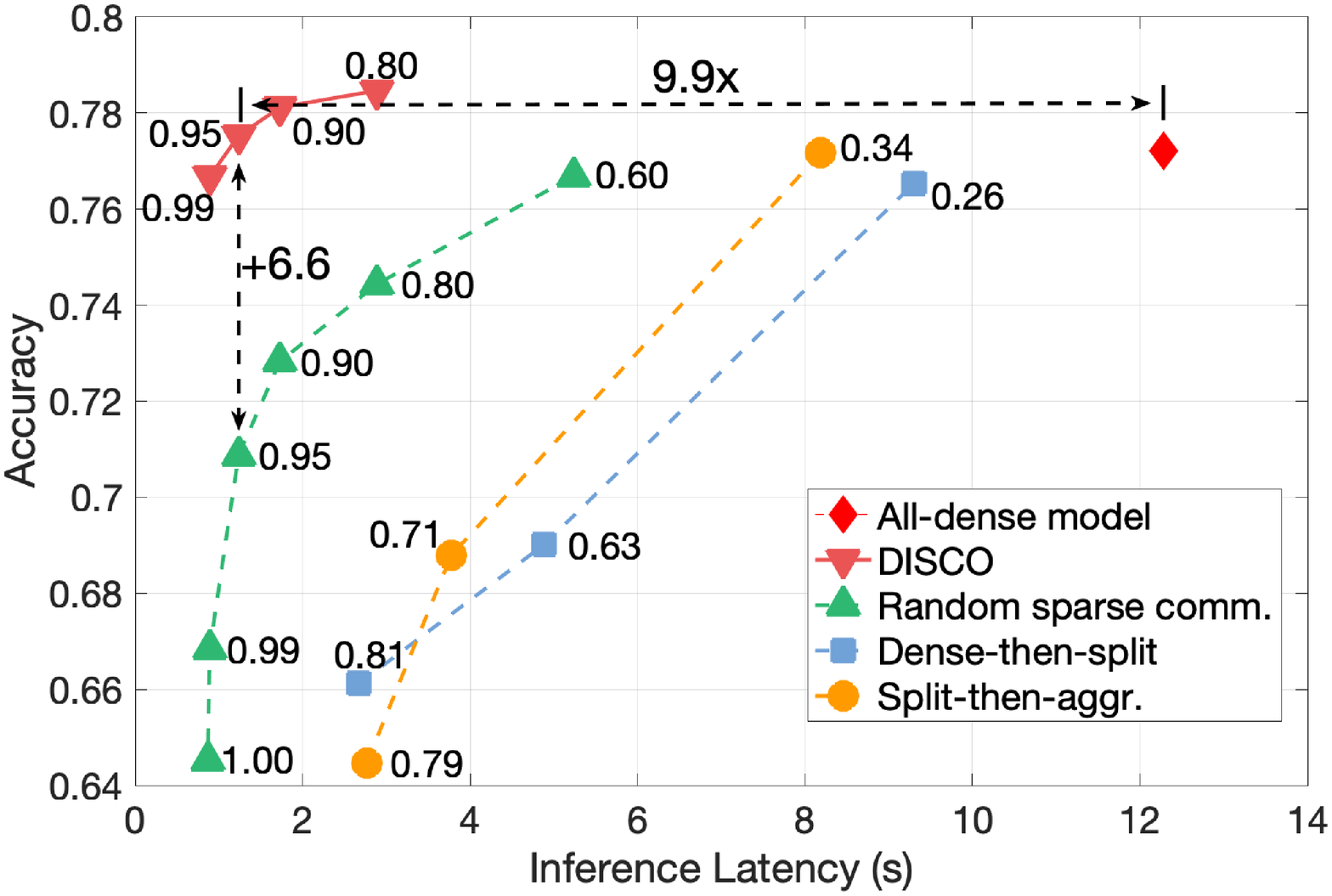}
  \caption{DeeplabV3+, PASCAL VOC2012, mIoU}
  \label{fig:result_deeplabv3+}
\end{minipage}

\begin{minipage}[c]{0.45\linewidth}
 \hspace{-1em}
  \includegraphics[width=1\linewidth]{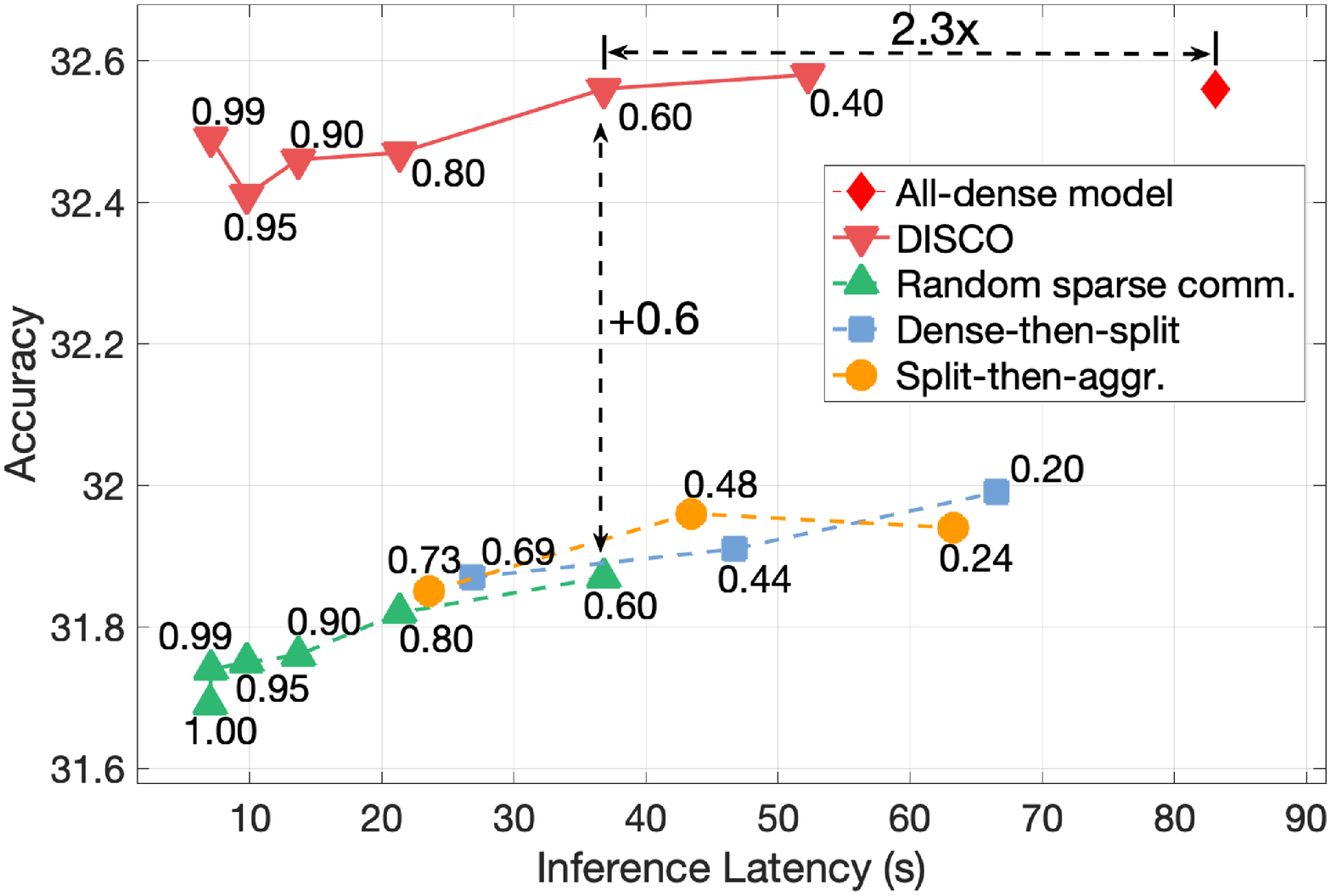}
  \caption{ESRGAN, DIV2K, PSNR}
  \label{fig:result_esrgan23}
\end{minipage}

\end{figure*}

In Figures~\ref{fig:result_rn50},~\ref{fig:deit},~\ref{fig:result_ssd300},~\ref{fig:result_deeplabv3+},~\ref{fig:result_esrgan23}, we investigate the latency and the accuracy trade-offs of various DNN models, datasets, and evaluation metrics when they are distributed across two nodes.  
We also show the $S_{\textrm{comm}}$ of each DNN next to the plotted result. 
Results on the larger number of nodes are presented in Section~\ref{sec:ablation}.
Details on the training recipe are presented in the Appendix. We vary $S_{\textrm{comm}}$ (and thus $S_{\textrm{comp}}$ based on Eq.~\ref{eq:scomm-and-scomp}) in the DISCO framework to obtain a sequence of DNN models with different distributed inference latency and accuracy. The latency estimation is based on Eq.~\ref{eq:Lcomm}, Eq.~\ref{eq:Lcomp}, and Eq.~\ref{eq:pipeline} for the ``pipeline'' mode. 
The system configuration is based on~\cite{dong2022cvprsplitnets} where $B = 37.5$ MB/s and $C = 125$ GOP/s. The input feature is assumed to be represented as 4-byte floating point numbers. Results on other configurations with publicly available specifications, such as NVIDIA A100 with NVLink, T4 with PCIe, Xeon CPU with ethernet, Cortex M4 processor with wireless communications, are summarized in Table~\ref{tab:summary_pipeline}.

We compare our framework (DISCO) with three different existing or baseline approaches. 
\begin{enumerate}
    \item Random sparse communications: The transmitted input features are randomly selected instead of being selected based on the significance of weights in Section~\ref{sec:selection}. In particular, when $S_{\textrm{comm}}=1$, there is no communication between nodes and DNNs are partitioned into ``independent branches''~\cite{Hadidi2020LCP}.
    \item Dense-then-split: The features in the first few layers are densely communicated, and the remaining layers are all separate without feature communication (Figure~\ref{fig:disco}(b)). Different splitting points result in different accuracy-latency trade-off. This architecture is proposed in~\cite{kim2017icmlsplitnet}. 
    \item Split-then-aggregate: The first few layers are all separate without communicated features, and the remaining layers have dense feature communications (Figure~\ref{fig:disco}(c)). Different aggregation points result in different accuracy-latency trade-offs. This architecture is proposed in~\cite{dong2022cvprsplitnets}. 
\end{enumerate}

To quantify the benefits of DISCO, we show two metrics. The first metric is the \textbf{latency reduction} from the densely communicated model without accuracy loss (the horizontal dashed double arrow in Figures~\ref{fig:result_rn50},~\ref{fig:deit},~\ref{fig:result_ssd300},~\ref{fig:result_deeplabv3+},~\ref{fig:result_esrgan23}). 
The second metric is the \textbf{accuracy improvement} over all baseline and existing methods with the specified latency in the first metric (the vertical dashed double arrow in Figures~\ref{fig:result_rn50},~\ref{fig:deit},~\ref{fig:result_ssd300},~\ref{fig:result_deeplabv3+},~\ref{fig:result_esrgan23}). 

The observation that our results outperform the random sparse communications shows the effectiveness of DISCO in selecting features and training the DNN (Section~\ref{sec:selection},~\ref{sec:iterative_training}). On the other hand, our results outperform ``Dense-then-split'' and ``Split-then-aggregate'', which shows that the sparse feature communications between nodes should exist in most of the layers, instead of concentrating on a subset of layers and completely being overlooked by the rest of layers.

\subsection{ResNet50 models on image classification}
The ResNet architecture is widely used in computer vision tasks. We use ResNet-50 models on ImageNet~\cite{deng2009imagenet} 
to demonstrate the benefit of DISCO compared to existing and baseline methods. 


Figure~\ref{fig:result_rn50} shows that DISCO outperforms prior arts with the following advantages: 
1) The accuracy improvement is 1.6\%. 
2) The latency reduction is 4.7x (from 1.14s to 0.24s).
3) Compared to the model with indenpendent branches ($S_{\textrm{comm}}=100\%$), by adding only 1\% feature communications ($S_{\textrm{comm}}=99\%$), which slightly increases the inference latency from 32ms to 35ms, the accuracy can be improved by a large margin from 73.4\% to 75.9\%. This demonstrates that a small portion of communicated features can significantly increase the model accuracy.

\begin{table*}[t]
\small
\centering
\caption{A summary of latency and accuracy improvement by DISCO}
\scalebox{1}{
\begin{tabular}{c | c | c | c | c | c }
\toprule
Models & ResNet-50 & DeiT & SSD & DeepLabV3+ & ESRGAN \\ \hline
\begin{tabular} {c}
NVIDIA-T4 (65TFLOPs)\\ PCIe (32GB/s) 
\end{tabular}& 4.7x & 1.7x  & 8.5x  & 8.8x  & 2.3x \\ \hline
\begin{tabular} {c} NVIDIA-A100 (312TFLOPs)\\ NVLink (600GB/s)\end{tabular} & 4.2x & 1.7x & 5.0x & 4.5x & 2.6x\\ \hline
\begin{tabular} {c} Xeon CPU (32$\times$3GFLOPs)\\ Ethernet (125MB/s)\end{tabular} & 3.9x & 1.7x & 5.4x & 5.7x & 2.1x \\ \hline
\begin{tabular} {c} ARM Cortex-M4 (64MFLOPs)\\ Wireless (1Mbps)\end{tabular} & 4.2x & 1.7x & 5.0x & 4.4x & 2.3x \\ \hline
Accuracy Improv. & +1.6 & +1.6 & +2.8 & +6.6 & +0.6 \\
\bottomrule
\end{tabular}}
\label{tab:summary_pipeline}
\end{table*}

\subsection{DeiT models on image classification}
There is a growing interest in Vision Transformers~\cite{vit2020,deit2021} and we also apply the DISCO to DeiT-small models. 
We follow the same training recipe as in~\cite{rw2019timm}.
From Figure~\ref{fig:deit} we can observe: 1) DISCO has over 1\% accuracy advantage over all existing or baseline methods.
2) By adding 1\% of data communications between the two nodes, DISCO methods can improve the accuracy from 74.58\% to 78.70\%.

\subsection{SSD models on object detection}
The single-shot detection (SSD) model~\cite{liu2016ssd} is a fast one-phase object detection model. We use the COCO2017~\cite{coco2017} dataset and follow the training recipe in~\cite{NVIDIA_SSD}. 

From Figure~\ref{fig:result_ssd300}, we can observe the following: 
1) DISCO reduces inference latency by 9.2x (from 3.4s to 0.37s) with equal accuracy of 26\% mAP~\cite{NVIDIA_SSD}. 
2) DISCO has clear accuracy advantage (+2.8\% mAP at the same inference latency)  
3) By introducing 1\% feature communications, the accuracy can be significantly improved from 19.6\% to 24.6\% while the latency is slightly increased from 109ms to 116ms. 

\subsection{DeepLabV3+ models on semantic segmentation}
We use the standard DeepLabV3+ models with ResNet-101 as the backbone~\cite{deeplabv3plus2018} for semantic segmentation task. The dataset is PASCAL VOC2012~\cite{pascalvoc2012}. 
We follow the training recipe in~\cite{deeplabv3plus2018}.

Figure~\ref{fig:result_deeplabv3+} shows the following: 
1)  DISCO reduces the inference latency by 9.9x (from 12.3s to 1.24s) with slightly better accuracy (77.5\% vs. 77.21\%~\cite{deeplabv3plus2018}). 
2) DISCO has a clear accuracy advantage (+6.6\%) over existing or baseline methods.
3) By introducing 1\% feature communications, the accuracy can be improved from 64.5\% to 76.7\% while the latency is only increased from 0.87s to 0.89s.

\subsection{ESRGAN models on super resolution}
We use the ESRGAN~\cite{wang2018esrgan} with 23 residual-in-residual dense blocks (RRDB)~\cite{wang2018esrgan} architecture to demonstrate the DISCO on image super resolution (up-scaling the height and width each by 4 times). The dataset is DIV2K~\cite{Agustsson_2017_CVPR_Workshops}.

Figure~\ref{fig:result_esrgan23} shows 
1) There is a clear PSNR advantage (around 0.6dB) for DISCO.
2) DISCO reduces the inference latency by 2.3x (from 83s to 36s) at the same PSNR=32.56dB.


\subsection{Results summaries}
In addition to the system configuration to generate Figures~\ref{fig:result_rn50},~\ref{fig:deit},~\ref{fig:result_ssd300},~\ref{fig:result_deeplabv3+},~\ref{fig:result_esrgan23} ($B = 37.5$ MB/s and $C = 125$ GOP/s~\cite{dong2022cvprsplitnets}), Table~\ref{tab:summary_pipeline} shows the latency and accuracy improvement for other typical computation-communication system with publicly available specifications. When our proposed models and baseline models are inferred using these systems, the accuracy will not change but the latency will. Therefore, we show the accuracy improvement of each DNN model for all systems, and show the latency improvement of each DNN model for each system. 

The geometric mean of latency reduction obtained by DISCO is 3.5x on 20 model and system configuration pairs. It also has clear accuracy advantages over the baseline or existing designs.

\section{{Ablation Study}} \label{sec:ablation}

\subsection{Number of Nodes}
\begin{table*}[t]
\small
\centering
\caption{Comparison of independent branches and DISCO with different number of nodes $N$. System configuration: $B = 37.5$MB/s and $C = 3.75$GOP/s. }
\begin{tabular}{c | c | c | c | c}
\toprule
Method & $N$ & $S_{\textrm{comm}}$ & Lat.(ms) & Accu.(\%) \\
\hline
ResNet50/8 & 1 & n/a  & 96.2 & 55.55 \\ \hdashline
Independent  & 2 & 100\% & 96.4 & 59.47 \\
branches & 4 & 100\% & 96.6 & 61.08 \\
\cite{Hadidi2020LCP} & 8 & 100\% & 97.2 & 64.74 \\ \hdashline
\multirow{3}{*}{DISCO} & 2 & 99\% & 96.7 & 62.31 \\
 & 4 & 98.7\% & 98.2 & 67.59 \\
 & 8 & 98.9\% & 101.2 & 71.89 \\
\bottomrule
\end{tabular}
\label{tab:rn50-multiple-nodes}
\end{table*}

One of the motivations for distributing DNN inference to multiple nodes is to reduce the memory requirement of each node. To demonstrate the benefit of multiple nodes, we make the following assumption: 1) We denote ResNet50/8 the DNN with 1/8 of the layer width of a standard ResNet50 model. And we assume each node has a memory capacity that is slightly larger than the requirement of inferring ResNet50/8. 2). The system has comparable communication and computation latency, where we assume $B = 37.5$MB/s and $C = 3.75$GOP/s in this subsection. 

If we have $N$ nodes and the memory of each node restricts it to infer a ResNet50/8, but not a significantly larger model. We compare two cases. 1) They form $N$ independent branches~\cite{Hadidi2020LCP}. 2) Those $N$ nodes can have a tiny amount of communications ($S_{\textrm{comm}}\approx 99\%$) between any pair of them (DISCO), such that the memory requirement to handle the extra 1\% data does not exceed the memory capacity of each node.

Table~\ref{tab:rn50-multiple-nodes} shows the comparison where a few phenomena can be observed. 1) With less then 2\% latency overhead, multiple-node distributed inference can significantly improve the accuracy of single-node inference of a small ResNet50/8 model (from 55.55\% to 67.59\%). 2) DISCO has (2.84\%, 6.51\%, 7.15\%) accuracy advantage over $N=2,4,8$ independent branches of ResNet50/8's with small (0.2ms to 4ms) latency overhead, respectively.

\subsection{Equilibrium sparsity distribution}
In Table~\ref{tab:equilibrium_results}, we compare the distribution of communication sparsity by using a 1) uniform distribution and 2) non-uniform distribution at the equilibrium point shown in Section~\ref{sec:equilibrium}. Since the equilibrium sparsity for each layer guarantees an equal latency for communication and computation, they can perfectly hide each other. Therefore, at the same inference latency, the average sparsity of both communication and computation are smaller, leading to 0.74\% better top-1 accuracy of the ResNet-50 model. Note that for a fair comparison, both models are obtained by the DISCO framework directly from a densely communicated ResNet-50 model without iterative pruning method. Also note that $S_{\textrm{comm}}$ at the equilibrium is very high. This is because for a system with $B = 37.5$ MB/s and $C = 125$ GOP/s~\cite{dong2022cvprsplitnets}, the communication latency is much larger than the computation latency. If $B$ increases, $S_{\textrm{comm}}$ at the equilibrium can be smaller.

\begin{table}[t]
\small
\centering
\caption{Comparison of uniform and non-uniform $S_{\textrm{comm}}$ for ResNet-50 models. System configuration: $B = 37.5$MB/s and $C = 125$GOP/s.}
\scalebox{1}{
\begin{tabular}{c | c | c | c | c }
\toprule
$S_{\textrm{comm}}$ type & $S_{\textrm{comm}}$  & $S_{\textrm{comp}}$  &  Latency & Top-1 Accu  \\ \hline
Uniform  & 0.99 & 0.495 & 35.4$ms$ & 75.55\% \\
Equilibrium  & 0.97 & 0.46 & 32.9$ms$ & 76.29\% \\
\bottomrule
\end{tabular}}
\label{tab:equilibrium_results}
\end{table}

\section{Conclusions}
In this paper, we propose a framework, called DISCO, to train DNNs for distributed inference with sparse communications. We convert the problem of selecting the subset of data for transmission to a DNN sparsification problem and show that DISCO can reduce the latency by an average of 3.5x or improve the accuracy significantly. 

{\small
\bibliographystyle{ieee_fullname}
\bibliography{reference}

\begin{thebibliography}{10}\itemsep=-1pt

\bibitem{NestedDQ2019}
A. Abdi and Faramarz Fekri.
\newblock Nested dithered quantization for communication reduction in
  distributed training.
\newblock {\em arXiv preprint arXiv:1904.01197}, 2019.

\bibitem{AbrahamyanChenBekoulis2021}
Lusine Abrahamyan, Yiming Chen, Giannis Bekoulis, and Nikos Deligiannis.
\newblock Learned gradient compression for distributed deep learning.
\newblock {\em IEEE Transactions on Neural Networks and Learning Systems},
  2021.

\bibitem{Agustsson_2017_CVPR_Workshops}
Eirikur Agustsson and Radu Timofte.
\newblock Ntire 2017 challenge on single image super-resolution: Dataset and
  study.
\newblock In {\em The IEEE Conference on Computer Vision and Pattern
  Recognition (CVPR) Workshops}, July 2017.

\bibitem{Aji2017SparseCF}
Alham~Fikri Aji and Kenneth Heafield.
\newblock Sparse communication for distributed gradient descent.
\newblock In {\em EMNLP}, 2017.

\bibitem{QSGD2017}
Dan Alistarh, Demjan Grubic, Jerry Li, Ryota Tomioka, and Milan Vojnovic.
\newblock {QSGD}: Communication-efficient {SGD} via gradient quantization and
  encoding.
\newblock In {\em Advances in Neural Information Processing Systems},
  volume~30, 2017.

\bibitem{ChenYangCheng2018}
Chi-Chung Chen, Chia-Lin Yang, and Hsiang-Yun Cheng.
\newblock Efficient and robust parallel dnn training through model parallelism
  on multi-gpu platform.
\newblock {\em arXiv preprint arXiv:1809.02839}, 2018.

\bibitem{deeplabv3plus2018}
Liang-Chieh Chen, Yukun Zhu, George Papandreou, Florian Schroff, and Hartwig
  Adam.
\newblock Encoder-decoder with atrous separable convolution for semantic image
  segmentation.
\newblock In {\em Proceedings of the European conference on computer vision
  (ECCV)}, 2018.

\bibitem{ParallelismNIPS2012}
Jeffrey Dean, Greg Corrado, Rajat Monga, Kai Chen, Matthieu Devin, Mark Mao,
  Marc\textquotesingle~aurelio Ranzato, Andrew Senior, Paul Tucker, Ke Yang,
  Quoc Le, and Andrew Ng.
\newblock Large scale distributed deep networks.
\newblock In {\em Advances in Neural Information Processing Systems},
  volume~25, 2012.

\bibitem{deng2009imagenet}
Jia Deng, Wei Dong, Richard Socher, Li-Jia Li, Kai Li, and Li Fei-Fei.
\newblock {ImageNet}: A large-scale hierarchical image database.
\newblock In {\em Proceedings of the IEEE Conference on Computer Vision and
  Pattern Recognition (CVPR)}, pages 248--255, 2009.

\bibitem{dong2022cvprsplitnets}
Xin Dong, Ziyun Li, Meng Li, Zhongnan Qu, Barbara~De Salvo, Chiao Liu, and
  Hsiang-Tsung Kung.
\newblock Split{N}ets: Designing neural architectures for efficient distributed
  computing on head-mounted systems.
\newblock In {\em Proceedings of the IEEE Conference on Computer Vision and
  Pattern Recognition (CVPR)}, 2022.

\bibitem{vit2020}
Alexey Dosovitskiy, Lucas Beyer, Alexander Kolesnikov, Dirk Weissenborn,
  Xiaohua Zhai, Thomas Unterthiner, Mostafa Dehghani, Matthias Minderer, Georg
  Heigold, Sylvain Gelly, Jakob Uszkoreit, and Neil Houlsby.
\newblock An image is worth 16x16 words: Transformers for image recognition at
  scale.
\newblock {\em CoRR}, abs/2010.11929, 2020.

\bibitem{JointDNN2021}
Amir~Erfan Eshratifar, Mohammad~Saeed Abrishami, and Massoud Pedram.
\newblock {JointDNN}: An efficient training and inference engine for
  intelligent mobile cloud computing services.
\newblock {\em IEEE Transactions on Mobile Computing}, 20(2):565--576, 2021.

\bibitem{pascalvoc2012}
Mark Everingham and John Winn.
\newblock The {PASCAL} visual object classes challenge 2012 ({VOC2012})
  development kit.
\newblock {\em Pattern Analysis, Statistical Modelling and Computational
  Learning, Tech. Rep}, 8:5, 2011.

\bibitem{fekri2018towards}
Pedram Fekri, Peyman Setoodeh, Fariba Khosravian, AA Safavi, and Mehrdad~H
  Zadeh.
\newblock Towards deep secure tele-surgery.
\newblock In {\em Proceedings of the international conference on scientific
  computing (CSC)}, pages 81--86, 2018.

\bibitem{AMPNet2017}
Alexander~L. {Gaunt}, Matthew~A. {Johnson}, Maik {Riechert}, Daniel {Tarlow},
  Ryota {Tomioka}, Dimitrios {Vytiniotis}, and Sam {Webster}.
\newblock {AMPNet}: Asynchronous model-parallel training for dynamic neural
  networks.
\newblock {\em arXiv preprint arXiv:1705.09786}, 2017.

\bibitem{Giacomoni2008}
John Giacomoni, Tipp Moseley, and Manish Vachharajani.
\newblock Fastforward for efficient pipeline parallelism: A cache-optimized
  concurrent lock-free queue.
\newblock page 43–52. Association for Computing Machinery, 2008.

\bibitem{Gordon2006ExploitingCT}
Michael~I. Gordon, William Thies, and Saman~P. Amarasinghe.
\newblock Exploiting coarse-grained task, data, and pipeline parallelism in
  stream programs.
\newblock In {\em ASPLOS XII}, 2006.

\bibitem{grigorescu2020survey}
Sorin Grigorescu, Bogdan Trasnea, Tiberiu Cocias, and Gigel Macesanu.
\newblock A survey of deep learning techniques for autonomous driving.
\newblock {\em Journal of Field Robotics}, 37(3):362--386, 2020.

\bibitem{Guan2019XPipeEP}
Lei Guan, Wotao Yin, Dongsheng Li, and Xicheng Lu.
\newblock {XPipe}: Efficient pipeline model parallelism for multi-{GPU} {DNN}
  training.
\newblock {\em arXiv preprint arXiv:1911.04610}, 2019.

\bibitem{Hadidi2020LCP}
Ramyad Hadidi, Bahar Asgari, Jiashen Cao, Younmin Bae, Da~Eun Shim, Hyojong
  Kim, Sung-Kyu Lim, Michael~S. Ryoo, and Hyesoon Kim.
\newblock {LCP}: A low-communication parallelization method for fast neural
  network inference in image recognition.
\newblock {\em arXiv preprint arXiv:2003.06464}, 2020.

\bibitem{MCDNN2016}
Seungyeop Han, Haichen Shen, Matthai Philipose, Sharad Agarwal, Alec Wolman,
  and Arvind Krishnamurthy.
\newblock {MCDNN}: An approximation-based execution framework for deep stream
  processing under resource constraints.
\newblock In {\em Proceedings of the 14th Annual International Conference on
  Mobile Systems, Applications, and Services}, page 123–136. Association for
  Computing Machinery, 2016.

\bibitem{Hanhirova2018}
Jussi Hanhirova, Teemu {Kämäräinen}, Sipi Seppälä, Matti Siekkinen, Vesa
  Hirvisalo, and Antti Ylä-Jääski.
\newblock Latency and throughput characterization of convolutional neural
  networks for mobile computer vision.
\newblock {\em arXiv preprint arXiv:1803.09492}, 2018.

\bibitem{PipeDream2018}
Aaron Harlap, Deepak Narayanan, Amar Phanishayee, Vivek Seshadri, Nikhil
  Devanur, Greg Ganger, and Phil Gibbons.
\newblock Pipedream: Fast and efficient pipeline parallel {DNN} training.
\newblock {\em arXiv preprint arXiv:1806.03377}, 2018.

\bibitem{he2016deep}
Kaiming He, Xiangyu Zhang, Shaoqing Ren, and Jian Sun.
\newblock Deep residual learning for image recognition.
\newblock In {\em Proceedings of the IEEE conference on Computer Vision and
  Pattern Recognition (CVPR)}, 2016.

\bibitem{DistrEdge2022}
Xueyu Hou, Yongjie Guan, Tao Han, and Ning Zhang.
\newblock Distredge: Speeding up convolutional neural network inference on
  distributed edge devices.
\newblock In {\em 2022 IEEE International Parallel and Distributed Processing
  Symposium (IPDPS)}, pages 1097--1107, 2022.

\bibitem{huchenghao2022}
Chenghao Hu and Baochun Li.
\newblock Distributed inference with deep learning models across heterogeneous
  edge devices.
\newblock In {\em IEEE INFOCOM 2022 - IEEE Conference on Computer
  Communications}, pages 330--339, 2022.

\bibitem{HuDiyi2020}
Diyi Hu and Bhaskar Krishnamachari.
\newblock Fast and accurate streaming cnn inference via communication
  compression on the edge.
\newblock In {\em 2020 IEEE/ACM Fifth International Conference on
  Internet-of-Things Design and Implementation (IoTDI)}, pages 157--163, 2020.

\bibitem{Offloading2018}
Hyuk-Jin Jeong, InChang Jeong, Hyeon-Jae Lee, and Soo-Mook Moon.
\newblock Computation offloading for machine learning web apps in the edge
  server environment.
\newblock In {\em 2018 IEEE 38th International Conference on Distributed
  Computing Systems (ICDCS)}, pages 1492--1499, 2018.

\bibitem{JiaLin2018}
Zhihao Jia, Sina Lin, Charles~R. Qi, and Alex Aiken.
\newblock Exploring hidden dimensions in parallelizing convolutional neural
  networks.
\newblock In {\em Proceedings of the 35th International Conference on Machine
  Learning, {ICML} 2018}, volume~80, pages 2279--2288, 2018.

\bibitem{Jung-Lee2021}
Sehun Jung and Hyang-Won Lee.
\newblock Optimization framework for splitting {DNN} inference jobs over
  computing networks, 2021.

\bibitem{Neurosurgeon2017}
Yiping Kang, Johann Hauswald, Cao Gao, Austin Rovinski, Trevor Mudge, Jason
  Mars, and Lingjia Tang.
\newblock Neurosurgeon: Collaborative intelligence between the cloud and mobile
  edge.
\newblock {\em ACM SIGARCH Computer Architecture News}, 45:615--629, 04 2017.

\bibitem{kim2017icmlsplitnet}
Juyong Kim, Yookoon Park, Gunhee Kim, and Sung~Ju Hwang.
\newblock {S}plit{N}et: Learning to semantically split deep networks for
  parameter reduction and model parallelization.
\newblock In {\em Proceedings of the 34th International Conference on Machine
  Learning}, pages 1866--1874, 2017.

\bibitem{SurveyDNN2021}
R.S. Latha, G.~R. R.~Sreekanth, R.C. Suganthe, and R.~Esakki Selvaraj.
\newblock A survey on the applications of deep neural networks.
\newblock In {\em 2021 International Conference on Computer Communication and
  Informatics (ICCCI)}, 2021.

\bibitem{DISSEC2022}
Qiang Li, Liang Huang, Zhao Tong, Ting-Ting Du, Jin Zhang, and Sheng-Chun Wang.
\newblock Dissec: A distributed deep neural network inference scheduling
  strategy for edge clusters.
\newblock {\em Neurocomputing}, 500:449--460, 2022.

\bibitem{LI201695}
X. Li, G. Zhang, K. Li, and W. Zheng.
\newblock Deep learning and its parallelization.
\newblock In Rajkumar Buyya, Rodrigo~N. Calheiros, and Amir~Vahid Dastjerdi,
  editors, {\em Big Data}, pages 95--118. 2016.

\bibitem{coco2017}
Tsung-Yi Lin, Michael Maire, Serge Belongie, Lubomir Bourdev, Ross Girshick,
  Pietro~Perona James~Hays, Deva Ramanan, C.~Lawrence Zitnick, and Piotr
  {Dollár}.
\newblock {COCO} 2017.

\bibitem{liu2016ssd}
Wei Liu, Dragomir Anguelov, Dumitru Erhan, Christian Szegedy, Scott Reed,
  Cheng-Yang Fu, and Alexander~C Berg.
\newblock {SSD}: Single shot multibox detector.
\newblock In {\em Proceedings of the European conference on computer vision
  (ECCV)}, pages 21--37. Springer, 2016.

\bibitem{DistilledSplit2019}
Yoshitomo Matsubara, Sabur Baidya, Davide Callegaro, Marco Levorato, and Sameer
  Singh.
\newblock Distilled split deep neural networks for edge-assisted real-time
  systems.
\newblock In {\em Proceedings of the 2019 Workshop on Hot Topics in Video
  Analytics and Intelligent Edges}, page 21–26. Association for Computing
  Machinery, 2019.

\bibitem{Mohammed2020}
Thaha Mohammed, Carlee Joe-Wong, Rohit Babbar, and Mario~Di Francesco.
\newblock Distributed inference acceleration with adaptive {DNN} partitioning
  and offloading.
\newblock In {\em IEEE INFOCOM 2020 - IEEE Conference on Computer
  Communications}, pages 854--863, 2020.

\bibitem{NVIDIA_SSD}
NVIDIA.
\newblock {SSD300 v1.1 For PyTorch}.
\newblock
  \url{https://github.com/NVIDIA/DeepLearningExamples/tree/master/PyTorch/Detection/SSD},
  2020.

\bibitem{CRIME2021}
Daniele~Jahier Pagliari, Roberta Chiaro, Enrico Macii, and Massimo Poncino.
\newblock Crime: Input-dependent collaborative inference for recurrent neural
  networks.
\newblock {\em IEEE Transactions on Computers}, 70(10):1626--1639, 2021.

\bibitem{pang2021deep}
Guansong Pang, Chunhua Shen, Longbing Cao, and Anton Van~Den Hengel.
\newblock Deep learning for anomaly detection: A review.
\newblock {\em ACM Computing Surveys (CSUR)}, 54(2):1--38, 2021.

\bibitem{DEFER2022}
Arjun Parthasarathy and Bhaskar Krishnamachari.
\newblock Defer: Distributed edge inference for deep neural networks.
\newblock In {\em 2022 14th International Conference on COMmunication Systems
  and NETworkS (COMSNETS)}, pages 749--753, 2022.

\bibitem{Megatron2019}
Mohammad Shoeybi, Mostofa Patwary, Raul Puri, Patrick LeGresley, Jared Casper,
  and Bryan Catanzaro.
\newblock Megatron-lm: Training multi-billion parameter language models using
  model parallelism.
\newblock {\em arXiv preprint arXiv:1909.08053}, 2019.

\bibitem{Deeperthings2021}
Rafael Stahl, Alexander Hoffman, Daniel Mueller-Gritschneder, Andreas
  Gerstlauer, and Ulf Schlichtmann.
\newblock Deeperthings: Fully distributed {CNN} inference on
  resource-constrained edge devices.
\newblock {\em International Journal of Parallel Programming}, 2021.

\bibitem{Strom2015}
Nikko {Ström}.
\newblock Scalable distributed {DNN} training using commodity gpu cloud
  computing.
\newblock In {\em Interspeech 2015}, 2015.

\bibitem{tan2019efficientnet}
Mingxing Tan and Quoc Le.
\newblock Efficientnet: Rethinking model scaling for convolutional neural
  networks.
\newblock In {\em Proceedings of the International Conference on Machine
  Learning (ICML)}, pages 6105--6114, 2019.

\bibitem{Tanaka2021AutomaticGP}
Masahiro Tanaka, Kenjiro Taura, Toshihiro Hanawa, and Kentaro Torisawa.
\newblock Automatic graph partitioning for very large-scale deep learning.
\newblock {\em 2021 IEEE International Parallel and Distributed Processing
  Symposium (IPDPS)}, pages 1004--1013, 2021.

\bibitem{DDDC2017}
Surat Teerapittayanon, Bradley McDanel, and H.T. Kung.
\newblock Distributed deep neural networks over the cloud, the edge and end
  devices.
\newblock In {\em 2017 IEEE 37th International Conference on Distributed
  Computing Systems (ICDCS)}, pages 328--339, 2017.

\bibitem{deit2021}
Hugo Touvron, Matthieu Cord, Matthijs Douze, Francisco Massa, Alexandre
  Sablayrolles, and Herve Jegou.
\newblock Training data-efficient image transformers and distillation through
  attention.
\newblock In {\em International Conference on Machine Learning}, volume 139,
  pages 10347--10357, July 2021.

\bibitem{WangHuangLi2019}
Minjie Wang, Chien-chin Huang, and Jinyang Li.
\newblock Supporting very large models using automatic dataflow graph
  partitioning.
\newblock In {\em Proceedings of the Fourteenth EuroSys Conference 2019}.
  Association for Computing Machinery, 2019.

\bibitem{wang2018esrgan}
Xintao Wang, Ke Yu, Shixiang Wu, Jinjin Gu, Yihao Liu, Chao Dong, Yu Qiao, and
  Chen~Change Loy.
\newblock {ESRGAN}: Enhanced super-resolution generative adversarial networks.
\newblock In {\em The European Conference on Computer Vision Workshops
  (ECCVW)}, September 2018.

\bibitem{rw2019timm}
Ross Wightman.
\newblock Pytorch image models.
\newblock \url{https://github.com/rwightman/pytorch-image-models}, 2019.

\bibitem{wikiesp32}
{Wikipedia contributors}.
\newblock {ESP32} --- {W}ikipedia{,} the free encyclopedia, 2022.
\newblock [Online; accessed 6-Feb-2022].

\bibitem{DeepThings2018}
Zhuoran Zhao, Kamyar~Mirzazad Barijough, and Andreas Gerstlauer.
\newblock Deepthings: Distributed adaptive deep learning inference on
  resource-constrained iot edge clusters.
\newblock {\em IEEE Transactions on Computer-Aided Design of Integrated
  Circuits and Systems}, 37(11):2348--2359, 2018.

\end{thebibliography}
}
\appendix
\newcommand{\RR}{{\mathbb{R}}}
\newcommand{\EE}{{\mathbb{E}}}
\allowdisplaybreaks

\setcounter{equation}{5}
\setcounter{table}{3}
\setcounter{figure}{10}

\section{Relating $S_{\textrm{comm}} $ and $S_{\textrm{comp}}$.} \label{sec:app_Scomm_Scomp}

From Figure~1(f) in the main context, we can observe that sparse communication can also bring computation reduction.
Let $(W,H,I,O)$ denote the shape of the convolution tensor of one dense layer with dense communications.
When this dense layer is evenly distributed across $N$ nodes, the convolution tensor in each node will be of shape $(W,H,I, \frac{O}{N})$. Among $I$ input features, $\frac{I}{N}$ of them are from local nodes and the other $\frac{(N-1)I}{N}$ is transferred from other nodes. If we reduce the number of transferred features to $I'$, then the communication reduction is
\begin{equation}\label{eq:scomm}
S_{\textrm{comm}} = 1 - \frac{I'}{\frac{(N-1)I}{N}} = 1 - \frac{I' N}{(N-1)I} = \frac{NI - I - I'N}{(N-1)I}.
\end{equation}
On the other hand, by transferring $I'$ features, the node has $\frac{I}{N} + I'$ features as the input to its convolution tensor, and  the shape of the convolution tensor is therefore $(W, H, \frac{I}{N} + I', \frac{O}{N})$. We can calculate the computation reduction of this layer as
\begin{equation}\label{eq:scomp}
S_{\textrm{comp}} = 1 - N\frac{(\frac{I}{N} + I')\frac{O}{N}}{IO} = 1 - \frac{I + I'N}{NI} = \frac{NI - I - I'N}{NI}.
\end{equation}
From Eq.~1 and Eq.~2 (in the main context) we can deduce that
\begin{equation} \label{eqapp:scomm-and-scomp}
S_{\textrm{comm}} = \frac{N}{N-1} S_{\textrm{comp}} .
\end{equation}
From Eq.~1, Eq.~2, and Eq.~\ref{eqapp:scomm-and-scomp}, we can calculate the latency of each components of the distributed inference for a given $S_{\textrm{comm}}$.

\section{Detailed explanation of experiments}
Figure 5, 6, 7, 8, 9 in the main context show curves and provide clearer visual improvement over baseline or existing designs. As an explanation to Table 1, we further list some numerical results with another system configuraiton of NVIDIA T4 GPU and PCIe link ($C=65$TOPs and $B=32$GB/s). The latency and accuracy improvement is similar to that of $C=125$GOPs and $B=37.5$MB/s in the main context.

\subsection{ResNet models on image classification}
The ResNet architecture~\cite{he2016deep} is widely used in computer vision tasks. We use ResNet-50 models to demonstrate the benefit of DISCO compared to existing and baseline methods.

All ResNet models are trained on ImageNet dataset~\cite{deng2009imagenet} for 150 epochs with 8 epochs for warm-up, batch size 256 per GPU, initial learning rate 0.256, cosine learning rate scheduler, momentum 0.875, weight decay 3e-5, mixup 0.2.
There are 4 stages in the ResNet-50 model. We choose the split and aggregate locations after the 1st, 2nd, and 3rd stages, resulting in 3 models for dense-then-split~\cite{kim2017icmlsplitnet}  and 3 models for split-then-aggregate~\cite{dong2022cvprsplitnets}, respectively. In this case, the communication and computation sparsities are 100\% for split layers and 0\% for non-split layers. Thus, we do not list the $S_\textrm{comm}$ and $S_\textrm{comp}$.

Figure~5 in the main context shows the inference latency and top-1 accuracy of ResNet-50 models when their inferences are distributed across two nodes. The detailed numerical results are summarized in Table~\ref{tab:rn50_detail}. It lists the communication sparsity ($S_\textrm{comm}$), computation sparsity ($S_\textrm{comp}$), inference latency calculated by T4 and PCIe specifications, and top-1 accuracy. Note that based on Eq.~\ref{eqapp:scomm-and-scomp}, $S_\textrm{comm}=2S_\textrm{comp}$ when $N=2$.

\begin{table*}[t]
\centering
\caption{Results of ResNet-50 models on ImageNet. }
\begin{tabular}{c | c | c | c | c | c}
\toprule
Method & $S_{\textrm{comm}}$ & $S_{\textrm{comp}}$ & Latency (ms) & Top-1 accu. (\%) & Comments\\ \hline
Dense~\cite{he2016deep}  & 0\% & 0\% & 1.34 & 76.80 \\  \hdashline
\multirow{4}{*}{ \begin{tabular}{c} Random Sparse \\ Communication (baseline) \end{tabular} } & 80\% & 40\% & 0.283 & 75.26 & \multirow{4}{*}{ \begin{tabular}{c} Uniform $S_{\textrm{comm}}$ \end{tabular} }\\
 & 90\% & 45\% & 0.154 & 74.90 \\
 & 95\% & 47.5\% & 0.0921 & 74.54 \\
 & 99\% & 49.5\% & 0.0521 & 73.85 \\ \hdashline
\multirow{3}{*}{ \begin{tabular}{c} Dense-then-split~\cite{kim2017icmlsplitnet} \end{tabular} }
 & - & - & 1.24 & 75.14 & Split @ Loc. 1 \\
& - & - & 0.927 & 74.55 & Split @ Loc. 2\\
& - & - & 0.446 & 73.3 & Split @ Loc. 3 \\ \hdashline
\multirow{3}{*}{ \begin{tabular}{c} Split-then-aggregate~\cite{dong2022cvprsplitnets} \end{tabular} }
 & - & - & 0.936 & 75.10 & Aggr. @ Loc. 1\\
 & - & - & 0.456 & 74.88 & Aggr. @ Loc. 2\\
& - & - & 0.147 & 73.66 & Aggr. @ Loc. 3 \\ \hdashline
\multirow{4}{*}{ \begin{tabular}{c} DISCO (Ours) \end{tabular} } & 80\% & 40\% & 0.283 & 76.84 & \multirow{4}{*}{ \begin{tabular}{c} Uniform $S_{\textrm{comm}}$ \end{tabular} }\\
 & 90\% & 45\% & 0.154 & 76.50 \\
 & 95\% & 47.5\% & 0.0921 & 76.26 \\
 & 99\% & 49.5\% & 0.0521 & 75.94 \\  \hdashline
Two independent branches (DISCO) & 100\% & 50\% & 0.0497 & 75.26 & 750 epochs\\
Two independent branches (w/o DISCO) & 100\% & 50\% & 0.0497 & 73.36 & 1500 epochs\\
\bottomrule
\end{tabular}
\label{tab:rn50_detail}
\end{table*}

\begin{table*}[htbp]
\centering
\caption{Results of DeiT-S models on ImageNet. }
\begin{tabular}{c | c | c | c | c | c}
\toprule
Method & $S_{\textrm{comm}}$ & $S_{\textrm{comp}}$ & Latency (ms) & Top-1 accu. (\%) & Comments\\ \hline
Dense~\cite{deit2021} & 0\% & 0\% & 1.43 & 79.94 \\  \hdashline
\multirow{5}{*}{ \begin{tabular}{c} Random Sparse \\ Communication (baseline) \end{tabular} } & 40\% & 20\% & 1.03 & 79.17 & \multirow{5}{*}{ \begin{tabular}{c} Uniform $S_{\textrm{comm}}$ \end{tabular} }\\
 & 60\% & 30\% & 0.825 & 78.51\\
 & 80\% & 40\% & 0.621 & 78.45 \\
 & 90\% & 45\% & 0.519 & 78.24 \\
 & 95\% & 47.5\% & 0.473 & 75.65 \\
 \hdashline
\multirow{2}{*}{ \begin{tabular}{c} Dense-then-split~\cite{kim2017icmlsplitnet} \end{tabular} }
 & - & - & 1.11 & 78.39 & Split @ Loc. 1 \\
& - & - & 0.778 & 76.51 & Split @ Loc. 2\\  \hdashline
\multirow{2}{*}{ \begin{tabular}{c} Split-then-aggregate~\cite{dong2022cvprsplitnets} \end{tabular} }
 & - & - & 1.11 & 79.26 & Aggr. @ Loc. 1\\
 & - & - & 0.778 & 78.51 & Aggr. @ Loc. 2\\
 \hdashline
\multirow{6}{*}{ \begin{tabular}{c} DISCO (Ours) \end{tabular} } & 40\% & 20\% & 1.03 & 80.05 & \multirow{6}{*}{ \begin{tabular}{c} Uniform $S_{\textrm{comm}}$ \end{tabular} }\\
 & 60\% & 30\% & 0.825 & 80.08 \\
 & 80\% & 40\% & 0.621 & 79.50 \\
 & 90\% & 45\% & 0.519 & 79.21 \\
 & 95\% & 47.5\% & 0.473 & 79.11 \\
 & 99\% & 49.5\% & 0.449 & 78.70\\ \hdashline
 Two independent branches (DISCO) & 100\% & 50\% & 0.449 & 78.00 & \\
Two independent branches (w/o DISCO) & 100\% & 50\% & 0.449 & 74.57 & \\
\bottomrule
\end{tabular}
\label{tab:deit_detail}
\end{table*}

\begin{table*}[h]
\centering
\caption{Results of SSD300 models on COCO2017. }
\begin{tabular}{c | c | c | c | c | c}
\toprule
Method & $S_{\textrm{comm}}$ & $S_{\textrm{comp}}$ & Latency (ms) & mAP (\%) & Comments\\ \hline
Dense~\cite{NVIDIA_SSD} & 0\% & 0\% & 4.01 & 26.0 \\  \hdashline
\multirow{7}{*}{ \begin{tabular}{c} Random Sparse \\ Communication (baseline) \end{tabular} } & 20\% & 10\% & 3.22 & 25.4 & \multirow{7}{*}{ \begin{tabular}{c} Uniform $S_{\textrm{comm}}$ \end{tabular} }\\
 & 40\% & 20\% & 2.42 & 24.9\\
 & 60\% & 30\% & 1.63 & 24.3 \\
 & 80\% & 40\% & 0.836 & 23.6 \\
 & 90\% & 45\% & 0.471 & 23.2 \\
 & 95\% & 47.5\% & 0.302 & 22.6 \\
 \hdashline
\multirow{3}{*}{ \begin{tabular}{c} Dense-then-split~\cite{kim2017icmlsplitnet} \end{tabular} }
 & - & - & 3.36 & 24.7 & Split @ Loc. 1 \\
& - & - & 1.77 & 21.8 & Split @ Loc. 2\\
& - & - & 0.898 & 20.3 & Split @ Loc. 3\\  \hdashline
\multirow{3}{*}{ \begin{tabular}{c} Split-then-aggregate~\cite{dong2022cvprsplitnets} \end{tabular} }
 & - & - & 3.30 & 25.5 & Aggr. @ Loc. 1\\
 & - & - & 2.43 & 24.6 & Aggr. @ Loc. 2\\
 & - & - & 0.842 & 21.5 & Aggr. @ Loc. 3\\
 \hdashline
\multirow{4}{*}{ \begin{tabular}{c} DISCO (Ours) \end{tabular} } & 80\% & 40\% & 0.834 & 26.7 & \multirow{4}{*}{ \begin{tabular}{c} Uniform $S_{\textrm{comm}}$ \end{tabular} }\\
 & 90\% & 45\% & 0.471 & 26.1 \\
 & 95\% & 47.5\% & 0.302 & 25.6 \\
 & 99\% & 49.5\% & 0.193 & 24.6 \\
 \hdashline
 Two independent branches (DISCO) & 100\% & 50\% & 0.187 & 22.6 & \\
Two independent branches (w/o DISCO) & 100\% & 50\% & 0.187 & 19.6 & \\
\bottomrule
\end{tabular}
\label{tab:ssd_detail}
\end{table*}

DISCO has the following advantages:
1) The accuracy improvement is 1.6\%.
2) The latency reduction is 4.7x (from 1.3ms to 0.28ms).
3) Compared to the model with all-independent branches (No DISCO used) ($S_{\textrm{comm}}=100\%$), by adding only 1\% feature communications ($S_{\textrm{comm}}=99\%$), which slightly increases the inference latency from 0.0497ms to 0.0521ms, the accuracy can be improved by a large margin from 73.4\% to 75.9\%. This demonstrates that a small portion of communicated features can significantly increase model accuracy.
4) If the target architecture is 2 independent branches, then models trained by DISCO have a 1.9\% accuracy advantage over non-DISCO method, even though the non-DISCO method has twice number of training epochs. This is because DISCO gradually reduces the model size, which may have a smoother loss surface during optimization.

There is another observation by comparing split-then-aggregate~\cite{dong2022cvprsplitnets} and dense-then-split~\cite{kim2017icmlsplitnet}. That is,
split-then-aggregate~\cite{dong2022cvprsplitnets} has better performance than dense-then-split~\cite{kim2017icmlsplitnet}. This phenomenon is also observed in all other DNN models and tasks, inferring that dense communications are favored in the later layers.

\begin{table*}[htbp]
\centering
\caption{Results of DeeplabV3+ models on PASCAL VOC2012. }
\begin{tabular}{c | c | c | c | c | c}
\toprule
Method & $S_{\textrm{comm}}$ & $S_{\textrm{comp}}$ & Latency (ms) & mIoU (\%) & Comments\\ \hline
Dense~\cite{deeplabv3plus2018} & 0\% & 0\% & 14.4 & 77.2 \\  \hdashline
\multirow{5}{*}{ \begin{tabular}{c} Random Sparse \\ Communication (baseline) \end{tabular} } & 60\% & 30\% & 6.14 & 76.7 & \multirow{5}{*}{ \begin{tabular}{c} Uniform $S_{\textrm{comm}}$ \end{tabular} }\\
 & 80\% & 40\% & 3.41 & 74.4 \\
 & 90\% & 45\% & 2.19 & 72.8 \\
 & 95\% & 47.5\% & 1.64 & 70.9\\
 & 99\% & 49.5\% & 1.29 & 66.8 \\
 \hdashline
\multirow{3}{*}{ \begin{tabular}{c} Dense-then-split~\cite{kim2017icmlsplitnet} \end{tabular} }
 & - & - & 11.0 & 76.5 & Split @ Loc. 1 \\
& - & - & 5.94 & 69.0 & Split @ Loc. 2\\
& - & - & 3.37 & 66.1 & Split @ Loc. 3 \\ \hdashline
\multirow{3}{*}{ \begin{tabular}{c} Split-then-aggregate~\cite{dong2022cvprsplitnets} \end{tabular} }
 & - & - & 9.63 & 77.2 & Aggr. @ Loc. 1\\
 & - & - & 4.54 & 68.8 & Aggr. @ Loc. 2\\
& - & - & 3.41 & 64.5 & Aggr. @ Loc. 3 \\ \hdashline
\multirow{4}{*}{ \begin{tabular}{c} DISCO (Ours) \end{tabular} } & 80\% & 40\% & 3.41 & 78.5 & \multirow{4}{*}{ \begin{tabular}{c} Uniform $S_{\textrm{comm}}$ \end{tabular} }\\
 & 90\% & 45\% & 2.19 & 78.1 \\
 & 95\% & 47.5\% & 1.64 & 77.5 \\
 & 99\% & 49.5\% & 1.29 & 76.7 \\  \hdashline
Two independent branches (DISCO) & 100\% & 50\% &  1.26 & 75.6  & \\
Two independent branches (w/o DISCO) & 100\% & 50\% & 1.26  & 64.5 & \\
\bottomrule
\end{tabular}
\label{tab:deeplabv3_detail}
\end{table*}

\subsection{DeiT models on image classification}
\begin{figure*}[htbp]
  \includegraphics[width=\linewidth]{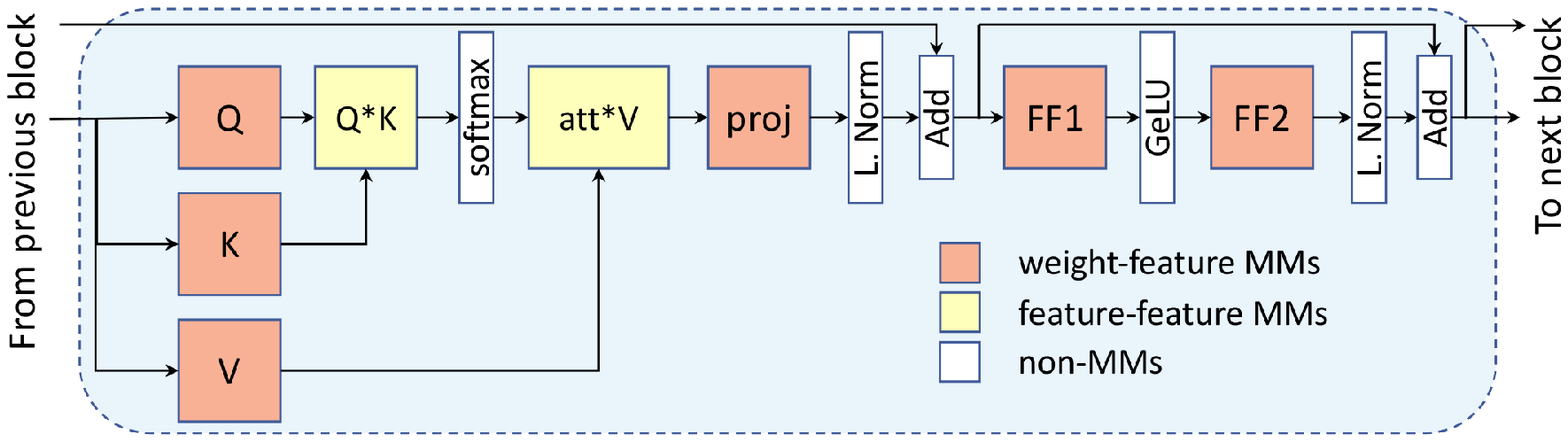}
  \caption{A Transformer block }
  \label{fig:transformer_block}
\end{figure*}
There is a growing interest in Vision Transformers~\cite{vit2020,deit2021} and we also apply the DISCO to DeiT-small models.
DeiT-small models have 12 blocks and there are eight layers of matrix-matrix multiplications within each block. Among those, six of them correspond to the multiplication between trainable weights and features (embeddings, activations, etc), which are denoted by Q, K, V, Proj, FF1, FF2 in Figure~\ref{fig:transformer_block}. The other two matrix multiplications are attention calculation between features, which are denoted by Q*K, and att*V. DISCO can be applied to the six weight-feature multiplications to reduce the communications during distributed inference. However, the other two feature-feature multiplications do not involve trainable weights, so that we assume the dense features are communicated between nodes for all cases in the experiments. Based on our profiling of the DeiT-small with dense communications, the six weight-feature multiplications contribute 75\% of overall data communication in a block, which can be reduced by DISCO.

There are 12 blocks in the DeiT-S model. We choose the split and aggregate locations after the 4th and 8th blocks, resulting in 2 models for dense-then-split~\cite{kim2017icmlsplitnet}  and 2 models for split-then-aggregate~\cite{dong2022cvprsplitnets}, respectively.

We follow the same training recipe as in~\cite{rw2019timm} where each DeiT model is trained for 300 epochs on ImageNet dataset, batch size 256 and learning rate 5e-4. The accuracy metric is the top-1 accuracy.
From Table~\ref{tab:deit_detail} we can observe that the DISCO has over 1\% accuracy advantage over the baseline design where features are randomly selected to be transmitted to the other nodes for inference. DISCO also has clear advantages over the dense-then-split~\cite{kim2017icmlsplitnet} and split-then-aggregate~\cite{dong2022cvprsplitnets} design.
Comparing the baseline model with two all-independent branches and our DISCO methods, we also observe that the accuracy is improved from 74.58\% to 78.70\% by introducing 1\% of data communication between the two nodes.

\subsection{SSD models on object detection}
The single-shot detection (SSD) model~\cite{liu2016ssd} is a fast one-phase object detection model. We use the COCO2017~\cite{coco2017} dataset and follow the training recipe in~\cite{NVIDIA_SSD}. The input image size is 300. Each data point in Table~\ref{tab:ssd_detail} is trained for 65 epochs with an initial learning rate 0.0026 and batch size 32. The learning rate is reduced by 10 times at the 43th and 54th epochs. The accuracy metric is the average mean precision (mAP).

SSD models 3 stages in the ResNet-50 backbone as one stage in the detection head. We choose the split and aggregate point after each stage, such that there are 3 models for dense-then-split and 3 models for split-then-aggregate, respectively.

From Table~\ref{tab:ssd_detail}, we can observe the following:
1) DISCO has clear advantages (around 3\% mAP at the same inference latency) over models with randomly selected features to communicate.
2) Compared to the densely communicated model, the sparsely communicated model obtained by DISCO can reduce the latency by 8.5 times (from 4ms to 0.47ms) with equal accuracy 26\%~\cite{NVIDIA_SSD}.
3) DISCO has significant improvement (5\% mAP) over the dense-then-split architecture~\cite{kim2017icmlsplitnet} and split-then-aggregate architecture~\cite{dong2022cvprsplitnets} design.
4) By introducing 1\% feature communications, the accuracy can be improved from 19.6\% to 24.6\% while the latency is only increased from 1.87ms to 1.93ms.

\subsection{DeepLabV3+ models on semantic segmentation}
We use the standard DeepLabV3+ models with ResNet-101 as the backbone~\cite{deeplabv3plus2018} for the semantic segmentation task.
The dataset is PASCAL VOC2012~\cite{pascalvoc2012}. We follow the training recipe in~\cite{deeplabv3plus2018} where each data point in Table~\ref{tab:deeplabv3_detail} is trained for 30k iterations with an initial learning rate 0.01, ``poly'' scheduler, batch size 16, crop size 513$\times$513 and output stride 16. The accuracy metric is the mean intersection-over-union (mIOU).
The split and aggregation points are after the 2nd, 3rd, and 4th stages in the backbone. Therefore, there are 3 models for dense-then-split~\cite{kim2017icmlsplitnet} and split-then-aggregate~\cite{dong2022cvprsplitnets} to compare.

From Table~\ref{tab:deeplabv3_detail}, we can observe the following:
1) DISCO has clear advantages (4\% to 10\% at the same inference latency) over models with random selection for sparsely communicated models.
2) Compared to the densely communicated model, the sparsely communicated model obtained by DISCO can reduce the latency by 8.8 times (from 14.4ms to 1.64ms) with slightly better accuracy (77.5\% for DISCO vs. 77.21\%~\cite{deeplabv3plus2018}).
3) DISCO has significant improvement over the dense-then-split architecture~\cite{kim2017icmlsplitnet} and split-then-aggregate architecture~\cite{dong2022cvprsplitnets} design.
4) By introducing 1\% feature communications, the accuracy can be improved from 64.5\% to 76.7\% while the latency is only increased from 1.26ms to 1.29ms.

\begin{table*}[htbp]
\centering
\caption{Results of ESRGAN models on DIV2K. }
\begin{tabular}{c | c | c | c | c | c}
\toprule
Method & $S_{\textrm{comm}}$ & $S_{\textrm{comp}}$ & Latency (ms) & PSNR (dB) & Comments\\ \hline
Dense~\cite{wang2018esrgan} & 0\% & 0\% & 97.4 & 32.56 \\  \hdashline
\multirow{5}{*}{ \begin{tabular}{c} Random Sparse \\ Communication (baseline) \end{tabular} } & 60\% & 30\% & 43.1 & 31.87 & \multirow{5}{*}{ \begin{tabular}{c} Uniform $S_{\textrm{comm}}$ \end{tabular} }\\
 & 80\% & 40\% & 25.0 & 31.82 \\
 & 90\% & 45\% & 16.0 & 31.76 \\
 & 95\% & 47.5\% & 11.5 & 31.75\\
 & 99\% & 49.5\% & 9.06 & 31.74 \\
 \hdashline
\multirow{3}{*}{ \begin{tabular}{c} Dense-then-split~\cite{kim2017icmlsplitnet} \end{tabular} }
 & - & - & 78.2 & 31.99 & Split @ Loc. 1 \\
& - & - & 55.1 & 31.91 & Split @ Loc. 2\\
& - & - & 32.1 & 31.87 & Split @ Loc. 3 \\ \hdashline
\multirow{3}{*}{ \begin{tabular}{c} Split-then-aggregate~\cite{dong2022cvprsplitnets} \end{tabular} }
 & - & - & 74.3 & 31.94 & Aggr. @ Loc. 1\\
 & - & - & 51.3 & 31.96 & Aggr. @ Loc. 2\\
& - & - & 28.2 & 31.85 & Aggr. @ Loc. 3 \\ \hdashline
\multirow{6}{*}{ \begin{tabular}{c} DISCO (Ours) \end{tabular} } & 40\% & 20\% & 61.2 & 32.58 & \multirow{6}{*}{ \begin{tabular}{c} Uniform $S_{\textrm{comm}}$ \end{tabular} }\\
 & 60\% & 30\% & 43.1 & 32.56 \\
 & 80\% & 40\% & 25.0 & 32.47 \\
 & 90\% & 45\% & 16.0 & 32.46 \\
 & 95\% & 47.5\% & 11.5 & 32.41 \\
 & 99\% & 49.5\% & 9.06 & 32.49 \\
\bottomrule
\end{tabular}
\label{tab:esrgan_detail}
\end{table*}

\subsection{ESRGAN models on super resolution}
We use the ESRGAN~\cite{wang2018esrgan} with 23 residual-in-residual dense blocks (RRDB) architecture to demonstrate the DISCO on image super resolution (up-scaling the height and width each by 4 times). The dataset is DIV2K~\cite{Agustsson_2017_CVPR_Workshops}. Each data point in Table~\ref{tab:esrgan_detail} is trained for 1008 epochs to guarantee the convergence. Following the training recipe in~\cite{wang2018esrgan}, the initial learning rate of the ADAM optimizer is set to 2e-4, the batch size is 16, and the spatial size of the cropped high-resolution patch is 128$\times$128. The accuracy metric is the peak-signal-to-noise-ratio (PSNR) between the ground-truth images and super-resolved images. The SSIM metric shows the same trend, thus not plotted.

The split and aggregation points are after the 6th, 12th, and 18th RRDB blocks and thus there are 3 models for dense-then-split~\cite{kim2017icmlsplitnet} and 3 models for split-then-aggregate~\cite{dong2022cvprsplitnets}, respectively.

We can draw some observations from Table~\ref{tab:esrgan_detail}.
1) There is a clear PSNR advantage (around 0.6dB) for DISCO over all baseline (randomly selecting communicated features) or existing architecture~\cite{kim2017icmlsplitnet,dong2022cvprsplitnets}.
2) The inference latency can be reduced by 2.26 times (from 97.4ms to 43.1ms) by DISCO compared to the densely communicated model at the same PSNR=32.56dB.

\begin{table*}[htbp]
\centering
\caption{Results of EfficientNet-B0 models on ImageNet. }
\begin{tabular}{c | c | c | c | c | c}
\toprule
Method & $S_{\textrm{comm}}$ & $S_{\textrm{comp}}$ & Latency (ms) & Top-1 (\%) & Comments\\ \hline
Dense~\cite{tan2019efficientnet} & 0\% & 0\% & 0.85 & 77.10 \\  \hdashline
\multirow{4}{*}{ \begin{tabular}{c} Random Sparse \\ Communication (baseline) \end{tabular} } & 40\% & 20\% & 0.70 & 76.42 & \multirow{4}{*}{ \begin{tabular}{c} Uniform $S_{\textrm{comm}}$ for PW conv. layers \end{tabular} }\\
 & 60\% & 30\% & 0.63 & 75.91 \\
 & 80\% & 40\% & 0.55 & 75.65 \\
 & 90\% & 45\% & 0.51 & 75.60\\
 \hdashline
\multirow{2}{*}{ \begin{tabular}{c} Dense-then-split~\cite{kim2017icmlsplitnet} \end{tabular} }
 & - & - & 0.81 & 76.02& Split @ Loc. 1 \\
& - & - & 0.73 & 74.69 & Split @ Loc. 2\\
 \hdashline
\multirow{2}{*}{ \begin{tabular}{c} Split-then-aggregate~\cite{dong2022cvprsplitnets} \end{tabular} }
 & - & - & 0.61 & 76.51 & Aggr. @ Loc. 1\\
 & - & - & 0.52 & 75.89 & Aggr. @ Loc. 2\\
\hdashline
\multirow{4}{*}{ \begin{tabular}{c} DISCO (Ours) \end{tabular} } & 40\% & 20\% & 0.70 & 77.07 & \multirow{4}{*}{ \begin{tabular}{c} Uniform $S_{\textrm{comm}}$ for PW conv. layers \end{tabular} }\\
 & 60\% & 30\% & 0.63 & 77.12 \\
 & 80\% & 40\% & 0.55 & 76.54\\
 & 90\% & 45\% & 0.51 & 76.10 \\
\bottomrule
\end{tabular}
\label{tab:efficientnetb0_detail}
\end{table*}

\subsection{EfficientNet models on image classification}
We also explore the benefits of DISCO on EfficientNet architectures~\cite{tan2019efficientnet}. EfficientNet architectures have two types of convolutions, point-wise (PW) $1\times1$ convolutions and depth-wise separable (DWS) convolutions. The DWS convolutions are intrinsically compatible with the DISCO, in that the number of output features of DWS convolution equals the number of input features, and each output is only dependent on one and only one input feature. Therefore, the DWS convolution layers can be distributed across nodes without any communications between nodes. In order to apply DISCO, we sparsify the communications for the PW convolution layers.

We use EfficientNet-b0 as an example. We follow the same training recipe in~\cite{rw2019timm}. We train the models for 450 epochs with an initial learning rate 0.048, batch size 384. Detailed parameters in the optimizer is the same as~\cite{rw2019timm}.

EfficientNet-b0 has 6 stages, and we choose the split and aggregation point after the 2nd and 4th stages. Therefore, there are 2 models for dense-then-split~\cite{kim2017icmlsplitnet} and 2 models for split-then-aggregate~\cite{dong2022cvprsplitnets}, respectively.

Although it is widely accepted that it is difficult to reduce the computation cost for EfficientNet models without impacting the accuracy, we can observe some benefits of DISCO. 1) It can reduce the latency by 1.35 times (from 0.85ms to 0.63ms) without accuracy loss. 2) It has 0.6-1\% accuracy advantage over existing or baseline approaches with the same latency.

\end{document}